\documentclass[journal]{IEEEtran}
\usepackage{cite}
\usepackage[pdftex]{graphics} % for arXiv
\usepackage{epsfig} % for postscript graphics files
\usepackage[subrefformat=parens]{subcaption}
\usepackage{amsmath}
\usepackage{amssymb}  % assumes amsmath package installed

\newcommand{\argmax}{\mathop{\rm arg~max}\limits}

\newtheorem{remark}{Remark}

\usepackage{algorithm}
\usepackage{algorithmic}
\usepackage{array}
\begin{document}
\title{Continuous Deep Q-Learning with Simulator for Stabilization of Uncertain Discrete-Time Systems}

\author{Junya Ikemoto and Toshimitsu Ushio% <-this % stops a space
\thanks{This work was partially supported by JST-ERATO HASUO Project Grant Number JPMJER1603, Japan and JST CREST Grant Number JPMJCR2012, Japan.}% <-this % stops a space
\thanks{J. Ikemoto and T. Ushio are with the Graduate School of Engineering Science, Osaka University, Toyonaka, Osaka, 560- 8531, Japan. {\tt\small ikemoto@hopf.sys.es.osaka-u.ac.jp, ushio@sys.es.osaka-u.ac.jp}}%
}

\maketitle

\begin{abstract}
Applications of reinforcement learning (RL) to stabilization problems of real systems are restricted since an  agent needs many experiences to learn an optimal policy and may determine dangerous actions during its exploration. If we know a mathematical model of a real system, a simulator is useful because it predicts behaviors of the real system using the mathematical model with a given system parameter vector. We can collect many experiences more efficiently than interactions with the real system. However, it is difficult to identify the system parameter vector accurately. If we have an identification error, experiences obtained by the simulator may degrade the performance of the learned policy. Thus, we propose a practical RL algorithm that consists of two stages. At the first stage, we choose multiple system parameter vectors. Then, we have a mathematical model for each system parameter vector, which is called a virtual system. We obtain optimal Q-functions for multiple virtual systems using the continuous deep Q-learning algorithm. At the second stage, we represent a Q-function for the real system by a linear approximated function whose basis functions are optimal Q-functions learned at the first stage. The agent learns the Q-function through interactions with the real system online. By numerical simulations, we show the usefulness of our proposed method.
\end{abstract}


% \fi
\IEEEpeerreviewmaketitle

\section{Introduction}
\it Reinforcement Learning \rm (RL) is a subfield of machine learning \cite{Sutton_RL,Csaba_RL}. The standard framework of RL consists of a learner, which is called an \it agent\rm, and everything outside the learner, which is called an \it environment\rm. The agent determines an action based on its policy to interact with the environment, gets a reward for the interaction, and updates its policy using past interactions so as to maximize the sum of obtained rewards. RL has been paid much attention to in various fields and many practical applications have been proposed \cite{RL_example1,RL_example2,RL_example3,RL_example4,RL_example5}. In the control field, it is known that RL is strongly connected with optimal control methods from a theoretical point of view. In optimal control methods, we solve a \it Hamilton-Jacobi-Bellman \rm (HJB) equation derived from a model of a  system in order to compute optimal control inputs. However, in general, it is difficult to solve the equation analytically for a nonlinear system even if we can identify the model of the system because the HJB equation is a nonlinear equation. Thus, in order to obtain an approximate solution of the HJB equation, the \it adaptive dynamic programing \rm (ADP) was proposed \cite{ADP_1,ADP_2}. Since we often use a neural network as a function approximator, this method is also called the \it neuro-dynamic programing \rm (NDP) \cite{NDP}. In the ADP, the controller (or agent) solves the optimal control problem using data obtained through interactions with the system just as RL that is developed in the computational intelligence community. There are many applications of the ADP or RL to various control problems \cite{ADP_example1,ADP_example2,ADP_example3,ADP_example4,ADP_example5,ADP_example6}. 

Moreover, RL with deep neural networks (DNNs), which is called \it Deep RL \rm (DRL), has been successful in complicated control problems \cite{DRL}. The \it deep Q-network \rm (DQN) algorithm proposed by Mnih \it et al.\rm\ is the most famous algorithm \cite{DQN}. The DQN algorithm has achieved impressive results in many Atari 2600 video games using pixels as inputs directly. Moreover, the DQN algorithm has been improved by various means \cite{DDQN,DuelingNN,PER,HER,RDQN}. However, the DQN algorithm cannot be directly applied to problems in the continuous action domain. Thus, Lillicrap \it et al.\rm\ proposed the \it deep deterministic policy gradient \rm (DDPG) algorithm \cite{DDPG}, which is based on the \it actor-critic \rm method. We use two types of DNNs, which are called the \it actor \rm network and the \it critic \rm network. The DDPG algorithm can solve complicated systems such as humanoid robots in the physical simulator \it MuJoCo. \rm In \cite{hDDPG}, Yang \it et al.\rm\ proposed a hierarchical DRL algorithm based on the DDPG algorithm. Furthermore, Gu \it et al\rm.\ proposed the deep Q-learning algorithm for the continuous action domain \cite{NAF}. This algorithm is simpler than the DDPG algorithm because we use a single DNN. The large number of practical applications of DRL have been proposed \cite{RL_example1}, \cite{DRL_example1,DRL_example2,DRL_example3,DRL_example4,DRL_example5,DRL_example6}.

Nevertheless, applications of DRL to stabilization problems of real systems are restricted because the agent needs many experiences in order to learn its optimal policy. In addition, it is difficult to select appropriate hyper parameters such as learning rates. We must select these parameters heuristically by a large number of trials. Furthermore, if we apply DRL to control of a safety critical physical system, the agent in an early learning stage may determine actions that cause damage to the system during its explorations. In order to extend the application range of RL or DRL in the real systems, if we can use a mathematical model of the real system, it is useful to pre-train the policy with a simulator. The simulator predicts behaviors of the real system using the mathematical model with a given system parameter vector. Using the simulator, we can collect many experiences more efficiently and safely than interactions with the real system. In general, however, it is difficult to identify the system parameter vector accurately. If we have an identification error, the experiences obtained by the simulator may degrade the performance of the learned policy. Thus, in this paper, we propose a practical RL algorithm with a simulator taking the identification error into consideration. 

\subsection{Contributions}
The main contribution is that we propose a practical RL algorithm using multiple deep Q-functions learned with a simulator. It is known that the agent can learn its policy for a complicated system using DNNs. However, we need a large number of experiences to learn the optimal policy. Thus, we use a simulator which predicts behaviors of the real system using a mathematical model with a given system parameter vector, which will be called a \it virtual system. \rm It is useful to collect many experiences efficiently. However, even if a mathematical model of the real system is enough accurate to predict its behavior, the identification error of the system parameter vector may degrade the control performance of the learned policy for the real system. For the problem, we propose a practical RL algorithm that consists of two stages. At the first stage, we choose multiple system parameter vectors from a premised set. We prepare multiple systems with these chosen system parameter vectors in the simulator and obtain an optimal Q-function for each virtual system using the continuous deep Q-learning algorithm \cite{NAF}. At the second stage, we approximate an optimal Q-function for the real system as an approximated linear function whose basis functions are pre-trained optimal Q-functions at the first stage. Additionally, our proposed method can be applied to a system whose system parameter vector varies slowly. To the best of our knowledges, a practical Q-learning algorithm with pre-trained optimal Q-function using the continuous deep Q-learning has not been studied.

\subsection{Related Works}
At the first stage of our proposed method, we use the continuous deep Q-learning algorithm \cite{NAF} to learn optimal Q-functions for virtual systems from experiences obtained by the simulator. This algorithm can solve problems in the continuous action domain as with DDPG \cite{DDPG}, \it trust region policy optimization \rm (TRPO) \cite{TRPO}, and \it proximal policy optimization \rm (PPO) \cite{PPO} and has an advantage that the greedy actions are determined based on the learned optimal Q-function analytically. As an applicable example of this algorithm, Gu \it et al. \rm proposed a robotic manipulation method with asynchronous off-policy updates \cite{Gu}. 

Moreover, our proposed method is related to methods with simulators \cite{Kim, Rusu,Tan}. This is one of the approaches to extend the application range of DRL in the real world. To fill a gap between a  simulated system and a real system is an important issue.

As alternatives to learn the optimal policy efficiently, model-based approaches are useful \cite{MBRL_survey}. In these approaches, the agent learns the model of the system and optimizes its policy based on the learned model in various ways, for example, the \it dynamic programing \rm \cite{Bellman_DP}, the \it iterative linear quadratic regulation \rm method \cite{iLQR,iLQG} and gradient-based policy search methods such as the \it probabilistic inference for learning control \rm (PILCO) \cite{PILCO}. An application of the Lyapunov function \cite{Nonlinear_System} to the learned model is also a useful approach \cite{Berkenkamp}. On the other hand, model-based approaches heavily depend on accuracy of the system model. If the agent cannot learn the accurate system model, the policy learned by the model-based approach may not perform well for the real system. Moreover, the simple model representation might lack in expressiveness for the real system. Thus, methods integrating model-free and model-based approaches have been proposed \cite{NAF,MBMFRL_1,MBMFRL_2}. In \cite{NAF}, Gu \it et al. \rm apply iLQG \cite{iLQG} based on the model learned by an iteratively refitted time-varying linear model \cite{Levine} in order to accelerate continuous deep Q-learning. In \cite{MBMFRL_1}, Nagabandi \it et al. \rm propose a DRL algorithm using a model predictive controller based on a learned DNN dynamics model to initialize the model-free learner. In \cite{MBMFRL_2}, Kurutach \it et al. \rm proposed the \it model-ensemble TRPO \rm algorithm. They use an ensemble of DNNs to reduce an effect of model bias. The agent collects experiences through interactions with learned DNN  models and learns its policy by the TRPO algorithm. In this paper, it is assumed that a mathematical model of the real system is known while an accurate system parameter vector is unknown. Thus, we use multiple virtual systems with premised system parameter vectors instead of learning the mathematical model. 

\subsection{Structure}
The paper is organized as follows. In Section II, we review the standard RL framework and Q-learning algorithms with approximated functions. In Section III, we formulate the problem. In Section IV, we propose a practical RL algorithm for this problem. In Section V, by numerical simulations, we demonstrate learning performances of our proposed method. In Section VI, we conclude this paper and show future works.  

%%%%%%%%%%%%%%%%%%%%%%%%%%%%%%%%%%%%%%%%%%%%%%%%%%%%%%%%%%%%%%%%%%%%%%%%%%%%%%%%
\section{PRELIMINARIES} 
This section reviews the standard framework of RL and Q-learning with approximated functions.   
\subsection{Reinforcement Learning}
RL is one of the machine learning methods \cite{Sutton_RL,Csaba_RL}. In RL, the learner is called an \it agent \rm and everything outside the learner is called an \it environment. \rm The agent learns its policy through interactions with the environment. At each discrete-time $k\in \mathbb{Z}_{\ge0}$, the agent observes the state of the environment $x[k]\in\mathcal{X}$ and determines the action $a[k]\in\mathcal{A}$ based on its policy $\mu$, where $\mathcal{X}$ and $\mathcal{A}$ are sets of environment's states and agent's actions, respectively. In this paper, we assume that the agent's policy is deterministic, that is, $\mu:\mathcal{X}\to\mathcal{A}$. At the next discrete-time $k+1$, the agent observes the next state $x[k+1]\in\mathcal{X}$ and the reward $r[k]\in\mathbb{R}$. The $k$-th transition of the environment is caused by the stochastic or deterministic dynamics $T$ that depends on $x[k]$ and $a[k]$. The $k$-th reward $r[k]$ is given by the following reward function $R:\mathcal{X}\times\mathcal{A}\to\mathbb{R}$.
\begin{eqnarray}
r[k]=R(x[k],a[k]).\label{reward}
\end{eqnarray}
A tuple $(x[k],a[k],x[k+1],r[k])$ obtained through an interaction with the environment is called by an \it experience\rm. The agent updates its policy $\mu$ using past experiences. 

The goal of RL is that the agent learns the policy that maximizes a discounted sum of rewards $\sum_{k=0}^{\infty}\gamma^{k}r[k]$, where $\gamma\in[0,1)$ is a \it discounted factor \rm to prevent its divergence. In RL, we define a \it value function \rm $V^{\mu}(x)$ and a \it Q-function \rm $Q^{\mu}(x,a)$  underlying the policy $\mu$ as follows:
\begin{eqnarray}
V^{\mu}(x)=E\left[\sum_{i=k}^{\infty}\gamma^{i-k}R(x[i],\mu(x[i]))\right],\ x[k]=x,\label{value}
\end{eqnarray}
\begin{eqnarray}
Q^{\mu}(x,a)=R(x,a)+E\left[\sum_{i=k+1}^{\infty}\gamma^{i-k}R(x[i],\mu(x[i]))\right],\nonumber\\
\ x[k]=x,\ a[k]=a.\label{Q_function}
\end{eqnarray}
In addition, we define an optimal value function and an optimal Q-function as follows:
\begin{eqnarray}
V^{*}(x)&=&\max_{\mu}V^{\mu}(x),\ \ \ \forall x\in\mathcal{X},\label{optimal_value_function}\\
Q^{*}(x,a)&=&\max_{\mu}Q^{\mu}(x,a),\ \ \ \forall x\in\mathcal{X},\ \forall a\in\mathcal{A}.
\end{eqnarray}
It is known that there exist optimal policies that share the same optimal Q-function.

The optimal Q-function satisfies the following equation.
\begin{eqnarray}
Q^{*}(x,a)=R(x,a)+\gamma\max_{a'\in\mathcal{A}}Q^{*}(x',a'),\ x'\sim T(\cdot|\ x,a).\label{optimal_Q_function}
\end{eqnarray}
In \it Q-learning\rm, the agent learns the optimal Q-function using a set of experiences obtained through interactions with the environment. The agent updates the optimal Q-function by reducing the following \it TD-error \rm  derived from Eq.\ (\ref{optimal_Q_function}).
\begin{eqnarray}
\delta[k]=t[k]-Q(x[k],a[k]),\label{TD_error}
\end{eqnarray}
where $t[k]=r[k]+\gamma \max_{a'\in\mathcal{A}}Q(x[k+1],a')$ is a target value. The agent has a \it Q-table \rm that consists of the Q-value $Q(x,a)$ for each tuple $(x,a)\in\mathcal{X}\times\mathcal{A}$ in the case where both $\mathcal{X}$ and $\mathcal{A}$ are finite sets. For example, when the agent obtains an experience $(x,a,x',r)$, it updates the value $Q(x,a)$ as follows:
\begin{eqnarray}
Q(x,a)\leftarrow Q(x,a)+\alpha\delta,
\end{eqnarray}
where $\alpha>0$ is a update rate and the TD-error $\delta$ is
\begin{eqnarray}
\delta=r+\gamma\max_{a'\in\mathcal{A}}Q(x',a')-Q(x,a).\nonumber
\end{eqnarray}
After learning the optimal Q-function, the agent greedily determines an action as follows:
\begin{eqnarray}
a[k]\in\argmax_{a\in\mathcal{A}}Q(x[k],a).\label{detrmine_action}
\end{eqnarray}  

\subsection{Q-learning with an approximated linear Q-function}
In the case where the states and/or the actions are continuous such as $\mathcal{X}=\mathbb{R}^{n_{x}}$ and $\mathcal{A}=\mathbb{R}^{n_{a}}$, we cannot make the Q-table and often parameterize the Q-function as follows:
\begin{eqnarray}
Q(x,a|w)=w^{\text{T}}\varphi(x,a),\label{app_Q_function}
\end{eqnarray}
where $w\in\mathbb{R}^{D}$ is a parameter vector and $\varphi=[\varphi_{1}\ \varphi_{2}\ ...\ \varphi_{D}]^{\text{T}}$ is a vector of basis functions $\varphi_{i}:\mathcal{X}\times\mathcal{A}\to\mathbb{R},\ i=1,2,...,D$.

The agent updates the parameter vector by
\begin{eqnarray}
w[k+1]&\leftarrow& w[k]+\alpha\delta[k]\frac{\partial Q(x[k],a[k]|w[k])}{\partial w}\nonumber\\
&=&w[k]+\alpha\delta[k]\varphi(x[k],a[k]),\label{update_param}
\end{eqnarray}
where $\alpha>0$ is a learning rate and $\delta[k]$ is a TD-error (\ref{TD_error}).

It is an important issue in this algorithm how to choose basis functions $\varphi_{i},\ i=1,2,...,D$. Additionally, since we must maximize the Q-function with respect to the action in order to compute the target value and determine the greedy action, it is desirable to choose basis functions that we can analytically maximize with respect to the action.  

\subsection{Continuous deep Q-learning}
In approximated Q-learning algorithms with DNNs, we need not choose basis functions beforehand. The continuous deep Q-learning algorithm \cite{NAF} is one of RL algorithms with DNNs. In order to describe the algorithm, we define the following \it advantage function \rm underlying the policy $\mu$.
\begin{eqnarray}
A^{\mu}(x,a):=Q^{\mu}(x,a)-V^{\mu}(x).\label{advantage_function}
\end{eqnarray}
This function represents the advantage of the action $a$ compared with the action $\mu(x)$. 

\begin{figure}
  \begin{center}
    \includegraphics[clip,width=8.5cm]{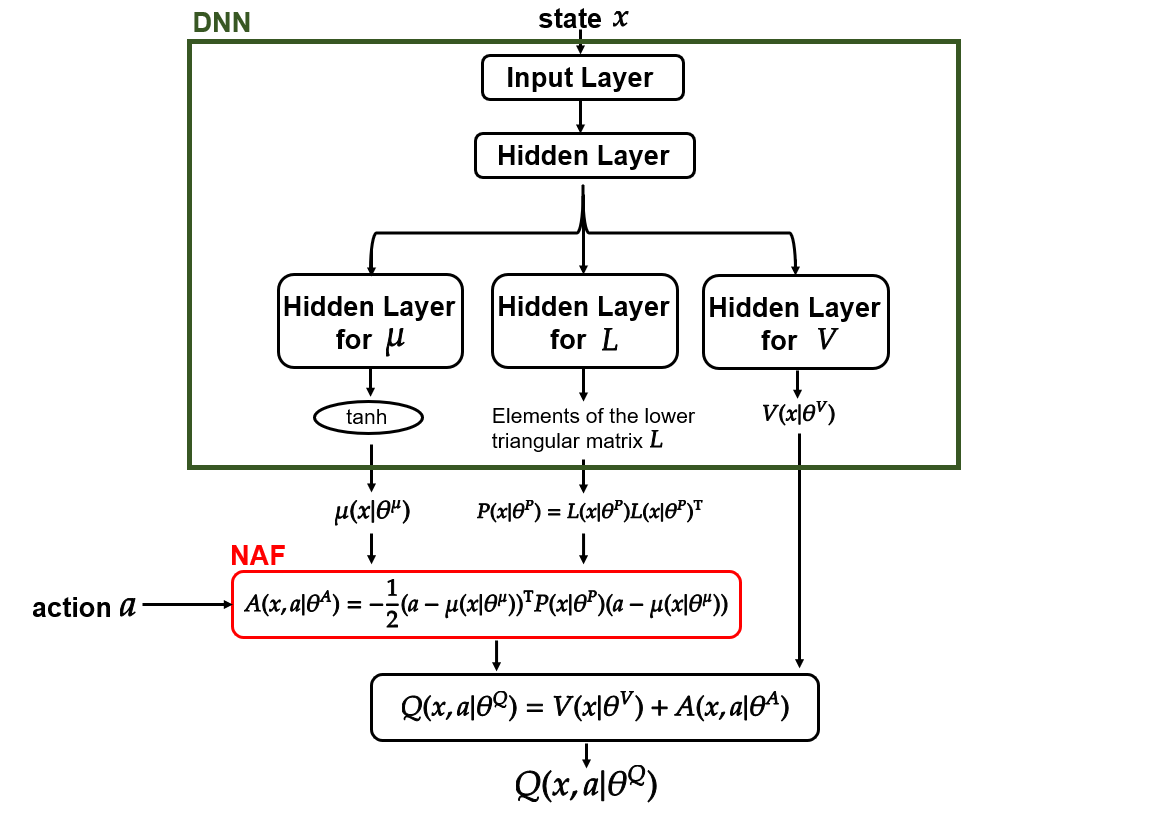}
    \caption{Illustration of the continuous deep Q-function that consists of a DNN and a quadratic function called a normalized advantage function (NAF). The NAF is quadratic with respect to the action and has a parameter matrix that is state-dependent and positive definite. }
    \label{NAF_NN}
  \end{center}
\end{figure}

We approximate the Q-function as shown in Fig.\ \ref{NAF_NN}. The DNN separately outputs a value function term $V$ and an advantage function term $A$ which is parameterized as a quadratic function of the action as follows:
\begin{eqnarray}
Q(x,a|\theta^{Q})&=&V(x|\theta^{V})+A(x,a|\theta^{A}),\label{NAF_Q}\\
A(x,a|\theta^{A})&=&-\frac{1}{2}(a-\mu(x|\theta^{\mu}))^{\text{T}}P(x|\theta^{P})(a-\mu(x|\theta^{\mu})),\nonumber\\
\label{NAF}
\end{eqnarray}
where $\theta^{V}$, $\theta^{\mu}$, and $\theta^{P}$ are parameter vectors for the value function term, the policy function term, and the parameter matrix of the advantage function, respectively. Let $\theta^{Q}=\{ \theta^{V},\theta^{\mu},\theta^{P}\}$ and $\theta^{A}=\{ \theta^{\mu}, \theta^{P} \}$. $P(x|\theta^{P})$ is a state-dependent matrix, which is given by $P(x|\theta^{P})=L(x|\theta^{P})L(x|\theta^{P})^{\text{T}}$, where $L(x|\theta^{P})$ is a lower-triangular matrix whose entries come from a linear output layer of the DNN. Its diagonal elements are set to be exponential. Hence, $P(x|\theta^{P})$ is a positive definite symmetric matrix. The advantage function term (\ref{NAF}) approximated by the quadratic form with respect to the actions is called a \it normalized advantage function \rm (NAF). The approximated Q-function (\ref{NAF_Q}) is more restrictive than a general DNN because the Q-function is quadratic with respect to the action $a$. On the other hand, we can analytically maximize the Q-function with respect to the action $a$. 

Shown in \bf Algorithm \ref{alg1} \rm is the continuous deep Q-learning algorithm with a NAF. From Line 1 to 2, we initialize parameter vectors of a \it main network \rm $\theta^{Q}$ and a \it target network \rm $\theta_{-}^{Q}$. At Line 3, we initialize a replay buffer $\mathcal{B}$. At Line 5, we initialize a random process or a probability distribution for exploration noises. At Line 6, the agent receives the initial state that is initialized randomly. From Line 8 to 10, the agent interacts with the environment and obtains an experience. In order to sample the experience, the agent determines an action by an exploration policy 
\begin{eqnarray}
\beta(x)=\mu(x|\theta^{\mu})+\epsilon,\label{exploration_policy}
\end{eqnarray} 
where the noise $\epsilon$ is generated from a random process or a probability distribution. After receiving the next state and the reward, the agent stores the experience in the replay buffer. From Line 11 to 14, the agent updates parameter vectors of DNNs, which is called the \it experience replay \rm \cite{DQN}. This technique leads to reductions of correlations between experiences. The agent selects $I$ experiences from the replay buffer randomly and updates the parameter vector $\theta^{Q}$ by minimizing the following error using these experiences $\{ (x^{(i)},a^{(i)},x'^{(i)},r^{(i)})\}_{i=1,...,I}$.
\begin{eqnarray}
L(\theta^{Q})=\sum_{i=1}^{I}\left(t^{(i)}-Q(x^{(i)},a^{(i)}|\theta^{Q})\right)^{2},\label{update_objection}
\end{eqnarray}
where 

\begin{eqnarray}
t^{(i)}&=&r^{(i)}+\gamma \max_{a'\in\mathcal{A}} Q(x'^{(i)},a'|\theta_{-}^{Q})\nonumber\\
&=&r^{(i)}+\gamma V(x'^{(i)}|\theta_{-}^{V}).\label{target_Q}
\end{eqnarray}
$\theta_{-}^{Q}$ and $\theta_{-}^{V}$ are parameter vectors of the target network. In order to make the learning more stable, $V(x'^{(i)}|\theta_{-}^{V})$ is outputted by the target network whose parameter vector is slowly updated by the soft update
\begin{eqnarray}
\theta_{-}^{Q}\leftarrow\tau\theta^{Q}+(1-\tau)\theta_{-}^{Q},\label{soft_update}
\end{eqnarray}
where $\tau>0$ is a very small positive constant. The technique is called the \it fixed target Q-network \rm \cite{DQN}. We minimize Eq.\ (\ref{update_objection}), which is a non-convex function so that we cannot optimize the parameter vector analytically. Thus, we often use \it stochastic gradient descent \rm (SGD) methods such as \it Adam \rm \cite{Adam}. 

%Fig
%algorithms
\begin{algorithm}[h]               
\caption{Continuous deep Q-learning with a NAF}         
\label{alg1}                          
\begin{algorithmic}[1]                  
\STATE Randomly initialize a parameter vector $\theta^{Q}$. 
\STATE Initialize the parameter vector of the target network $\theta^{Q}_{-}\leftarrow\theta^{Q}$.
\STATE Initialize the replay buffer $\mathcal{B}$.
\FOR{Episode $l=1,...,\text{MAX EPISODE}$}
\STATE Initialize a random process or a probability distribution for the exploration noise $\epsilon$.
\STATE Receive the initial state $x[0]$.
\FOR{Discrete-time step $k=0,...,K$}
\STATE Determine the action $a[k]=\mu(x[k]|\theta^{\mu})+\epsilon[k]$.
\STATE Execute $a[k]$ and receive the reward $r[k]$ and the next state $x[k+1]$.
\STATE Store the experience $(x[k],a[k],x[k+1],r[k])$ in the replay buffer $\mathcal{B}$.
\STATE Sample $I$ experiences $\{(x^{(i)},a^{(i)},x'^{(i)},r^{(i)}) \}_{i=1,...,I}$ from $\mathcal{B}$ randomly.
\STATE Set target values $t^{(i)}=r^{(i)}+\gamma V(x'^{(i)}|\theta^{V}_{-})$.
\STATE Update $\theta^{Q}$ by minimizing the loss: \\$L(\theta^{Q})=\frac{1}{I}\sum_{i=1}^{I}(t^{(i)}-Q(x^{(i)},a^{(i)}|\theta^{Q}_{-}))^{2}$.
\STATE Update the target network: $\theta_{-}^{Q}\leftarrow \tau\theta^{Q}+(1-\tau)\theta_{-}^{Q}$.
\ENDFOR
\ENDFOR
\end{algorithmic}
\end{algorithm}

%%%%%%%%%%%%%%%%%%%%%%%%%%%%%%%%%%%%%%%%%%%%%%%%%%%%%%%%%%%%%%%%%%%%%%%%%%%%%%%%

\section{CONTINUOUS DEEP Q-LEARNING USING SIMULATOR}
We consider the following discrete-time nonlinear deterministic system.
\begin{eqnarray}
x[k+1]=f(x[k],a[k]|\xi),\label{dynamics}
\end{eqnarray}
where $x\in\mathcal{X}(\subseteq\mathbb{R}^{n_{x}})$ and $a\in\mathcal{A}(\subseteq\mathbb{R}^{n_{a}})$ are the state and the control input of the system, respectively. It is assumed that we cannot identify the \it system parameter vector \rm $\xi=[\xi_{1}\ \xi_{2}\ ...\ \xi_{p}]^{\text{T}}\in\Xi\subset\mathbb{R}^{p}$ accurately, where $\Xi$ is a compact set of $\mathbb{R}^{p}$ and known beforehand. In this paper, we apply RL to stabilize the system (\ref{dynamics}) to the target state $x^{*}\in\mathcal{X}$ that is one of the fixed points of the system. We regard the system and the control input as the environment and the agent's action, respectively. The reward function is defined by 
\begin{eqnarray}
R(x,a)=-(x-x^{*})^{\text{T}}R_{1}(x-x^{*})-a^{\text{T}}R_{2}a,\label{reward_assumed}
\end{eqnarray}
where $R_{1}\in\mathbb{R}^{n_{x}\times n_{x}}$ and $R_{2}\in\mathbb{R}^{n_{a}\times n_{a}}$ are positive definite matrices. This is a standard reward function for stabilization and takes the maximum value $0$ at the target state $x^{*}$ with $a=0$.

If we have a mathematical model of the real system, a simulator is useful. The simulator predicts behaviors of the real system using the mathematical model with a given system parameter vector. We can  collect many experiences more efficiently than interactions with the real system. Thus, in this paper, we consider RL with the simulator. In general, however, we may have an identification error of the system parameter vector. Then, the experiences obtained by the simulator degrade the performance of the learned policy for the real system and, in the worst case, the policy may not stabilize the real system. In the next section, we propose a practical RL algorithm taking the identification error into account.

\section{Q-LEARNING WITH PRE-TRAINED MULTIPLE DEEP Q-NETWORKS}
\begin{figure*}[h]
  \begin{center}
    \includegraphics[clip,width=17.0cm]{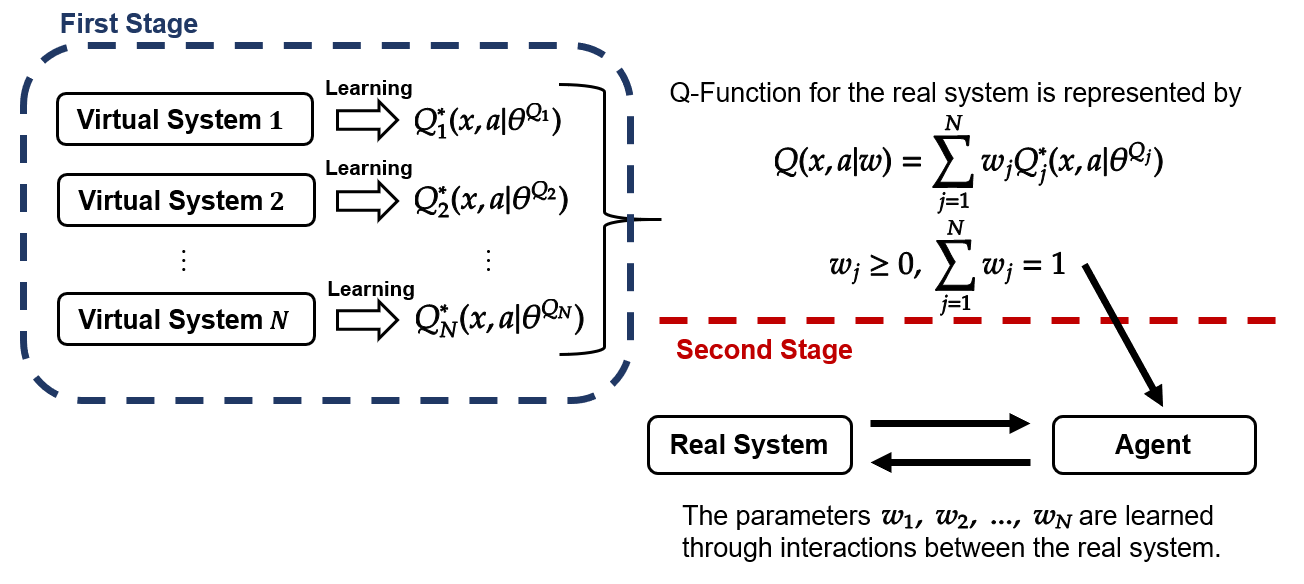}
    \caption{Illustration of our proposed method that consists of two stages. We choose $N$ system parameter vectors from $\Xi$ and prepare $N$ virtual systems. At the first stage, we obtain optimal Q-functions $Q_{j}^{*},\ j=1,2,...,N,$ for virtual systems using the continuous deep Q-learning algorithm. At the second stage, we represent the approximated Q-function for the real system with optimal Q-functions for virtual systems as basis functions. The agent learns the parameter vector $w$ through interaction with the real system.  }
    \label{Sim_Q}
  \end{center}
\end{figure*}

Although we can collect many experiences easily using a simulator, experiences obtained by the simulator may degrade the performance of the learned policy due to the identification error. To tackle this problem, we propose a practical RL algorithm that consists of two stages as shown in Fig.\ \ref{Sim_Q}. At the first stage, we choose $N$ system parameter vectors $\xi^{(j)},\ j=1,2,...,N$ from the premised system parameter set $\Xi$. Then, for each chosen parameter vector $\xi^{(j)}$, we have the mathematical model $f(x,a|\xi^{(j)})$, which will be called a virtual system (with $\xi^{(j)}$). Using the simulator, we collect experiences of the virtual system which are used for learning of an optimal Q-function and an optimal policy for the virtual system with $\xi^{(j)}$ by  the continuous deep Q-learning \cite{NAF}. At the second stage, we represent an optimal Q-function for the real system as an approximated linear Q-function whose basis functions are deep Q-networks learned for virtual systems at the first stage. The agent learns the parameter vector of the approximated linear Q-function through interactions with the real system.

\subsection{Q-Function for Real System}
At the first stage, we obtain the approximated optimal Q-function $Q_{j}^{*}(x,a|\theta^{Q_{j}})$ for each virtual system with $\xi^{(j)}$. Then, we approximate a Q-function for the real system as follows:
\begin{eqnarray}
Q(x,a|w)=\sum_{j=1}^{N}w_{j}Q_{j}^{*}(x,a|\theta^{Q_{j}}),\label{multi_Q_function}
\end{eqnarray}
where $w=\left[w_{1}\ w_{2}\ ...\ w_{N}\right]^{\text{T}}$ is a parameter vector of the Q-function. It is assumed that $\forall j\in\{1,2,...,N\},\ w_{j}\ge0$ and $\sum_{j=1}^{N}w_{j}=1$. In the approximated representation, basis functions $\varphi_{j}$ are optimal Q-functions $Q_{j}^{*}$ learned for the virtual systems. The agent learns the parameter vector $w$ through interactions with the real system as shown in Fig. \ref{Sim_Q}.

The optimal action $\mu(x|w)$ maximizes the Q-function $Q(x,a|w)$ with respect to the action $a$, that is,
\begin{eqnarray}
\mu(x|w)&=&\argmax_{a}Q(x,a|w)\nonumber\\
&=&\argmax_{a}\sum_{j=1}^{N}w_{j}A_{j}^{*}(x,a|\theta^{A_{j}})\label{n_NAF}.
\end{eqnarray}
In order to compute the action that maximizes the Q-function (\ref{multi_Q_function}), we solve 
\begin{eqnarray}
\frac{\partial}{\partial a}\sum_{j=1}^{N}w_{j}A_{j}^{*}(x,a|\theta^{A_{j}})=0.\label{optimal_u}
\end{eqnarray}
Then, we obtain the following stationary solution
\begin{eqnarray}
\hat{a}(x)=\sum_{j=1}^{N}\tilde{w}_{j}(x)\mu_{j}^{*}(x|\theta^{\mu_{j}}),\label{greedy_a}
\end{eqnarray}
where 
\begin{eqnarray}
\tilde{w}_{j}(x)=\left(\sum_{m=1}^{N}w_{m}P^{*}_{m}(x|\theta^{P_{m}})\right)^{-1}w_{j}P^{*}_{j}(x|\theta^{P_{j}}).\nonumber
\end{eqnarray}
See Appendix for its derivation. Note that $\sum_{j=1}^{N}w_{j}P^{*}_{j}(x|\theta^{P_{j}})$ is a positive definite symmetric matrix because parameter matrices of NAFs $P^{*}_{j}(x|\theta^{P_{j}}),\ j=1,...,N$ are positive definite symmetric matrices. Then, since the Hessian matrix
\begin{eqnarray}
 \frac{\partial^{2}}{\partial a^{2}}\sum_{j=1}^{N}w_{j}A_{j}^{*}(x,a|\theta^{A_{j}})=\sum_{j=1}^{N}w_{j}P_{j}^{*}(x|\theta^{P_{j}})\nonumber
\end{eqnarray} 
is positive definite, the stationary solution $\hat{a}(x)$ is the global optimal solution, that is, $\mu(x|w)=\hat{a}(x)$. 

\subsection{Q-Learning for Real System with Deep Q-Networks Learned for Multiple Virtual Systems}
The agent learns the parameter vector $w$ so as to reduce the TD-error using the standard Q-learning algorithm, where  parameters must satisfy the condition $\forall j\in\{1,2,...,N\},\ w_{j}\ge0$ and $\sum_{j=1}^{N}w_{j}=1$. 

At first, we introduce the following loss function that evaluates the TD-error for an experience $e=(x,a,x',r)$. 
\begin{eqnarray}
\mathcal{L}(w)=\frac{1}{2}\left(t-Q(x,a|w)\right)^{2},\label{based_td}
\end{eqnarray}
where 
\begin{eqnarray}
t=r+\gamma\max_{a'\in\mathcal{A}}Q(x',a'|w)\nonumber
\end{eqnarray} 
Note that we regard the target value $t$ as a constant value, where $a'$ is determined by Eq.\ (\ref{greedy_a}). From the first-order approximation at $w$, we have
\begin{eqnarray}
\mathcal{L}(w+\Delta w)-\mathcal{L}(w) \simeq \frac{\partial\mathcal{L}(w)}{\partial w}\Delta{w},\nonumber
\end{eqnarray}
where the Euclidean norm $\left|\left|\Delta w\right|\right|$ is small. We set $\Delta w=-\alpha\frac{\partial \mathcal{L}(w)}{\partial w}$, where $\alpha$ is a sufficiently small positive value such that $\mathcal{L}(w+\Delta w)-\mathcal{L}(w)<0$. Thus, we update $w$ based on the following rule to minimize the loss function (\ref{based_td}).
\begin{eqnarray}
w[k+1]\leftarrow w[k]-\alpha\frac{\partial \mathcal{L}(w[k])}{\partial w}.
\end{eqnarray}
This is equivalent to Eq. (\ref{update_param}). 
\begin{remark}
The update vector of the parameter vector $w$ is
\begin{eqnarray}
\alpha\frac{\partial \mathcal{L}(w)}{\partial w}=\alpha\delta \frac{\partial Q(x,a|w)}{\partial w}.\nonumber
\end{eqnarray}
Namely, the size of the update vector $\alpha\frac{\partial \mathcal{L}(w)}{\partial w}$ depends on the optimal Q-functions for the virtual systems. If we can obtain these optimal Q-functions, their outputs are close to 0 near the target state because of the reward function (\ref{reward_assumed}). Thus, the size of the update vector is small near the target state. 
\end{remark}

Next, we define the following loss function with a barrier term to keep all parameters nonnegative.
\begin{eqnarray}
\mathcal{L}^{B}(w)=\mathcal{L}(w)+\eta B(w), \label{loss}
\end{eqnarray}
where $\eta>0$ is a constant value and $B(w)$ is a barrier function. Let $\mathcal{W}=\{w\in\mathbb{R}^{N}|\ \forall n,\ w_{n}+\epsilon_{w}>0\}$, where $\epsilon_{w}>0$ is an arbitrarily small constant. The internal and the boundary of the set $\mathcal{W}$ are denoted by $\text{int}\mathcal{W}$ and $\partial\mathcal{W}$, respectively. The barrier function is given by
\begin{eqnarray}
B(w)\begin{cases}
	> 0 & w\in\text{int}\mathcal{W},\\
	\to+\infty & w\to \partial\mathcal{W}.
\end{cases}\label{barrier_function}
\end{eqnarray}
In our proposed method, we use the following update equation to learn the parameter vector $w$.
\begin{eqnarray}
w[k+1]&\leftarrow& w[k]-\alpha\frac{\partial \mathcal{L}^{B}(w[k])}{\partial w}\nonumber\\
&=&w[k] -\alpha\left( \frac{\partial \mathcal{L}(w[k])}{\partial w} + \eta\frac{\partial B(w[k])}{\partial w}\right).\nonumber\\
\label{w_update}
\end{eqnarray}

\begin{remark}
In our proposed algorithm, we reduce the update rate $\alpha$ in the case where there exist negative elements in the parameter vector updated by (\ref{w_update}).  We repeat reducing the learning rate by half until all elements of the updated parameter vector are nonnegative. 
\end{remark}

Shown in \bf Algorithms 2 \rm is our proposed learning algorithm with multiple deep Q-networks. The outline of \bf Algorithm 2 \rm is as follows. At Line\ 1, we choose $N$ system parameter vectors $\xi^{(j)},\ j=1,2,...,N$ from the premised set $\Xi$. At Line\ 2, we obtain Q-functions $Q_{j}^{*},\ j=1,2,...,N,$ for virtual systems using the continuous deep Q-learning as shown in \bf Algorithm 1\rm. At Line\ 3, we initialize the parameter vector $w$ and represent the Q-function for the real system. At Line\ 4, we initialize the state of the real system. From Line\ 5 to 20, the agent learns the parameter vector $w$ through interactions with the real systems online. From Line\ 12 to 18, if some elements of the updated parameter vector are negative after one update, we reduce the learning rate to keep all elements of the updated parameter vector nonnegative. At Line 19, we normalize the parameter vector after the update. By the algorithm, the agent learns its policy that stabilizes the real system to the target state $x^{*}$. 

\begin{algorithm}[h]               
\caption{Q-learning for the real system with multiple pre-trained deep Q-functions}         
\label{alg2}                          
\begin{algorithmic}[1] 
\STATE Choose $N$ system parameter vectors $\{\xi^{(j)}\}_{j=1,2,...,N}$.       
\STATE Obtain the $N$ optimal Q-functions $\{Q_{j}^{*}\}_{j=1,...,N}$ for virtual systems by \bf Algorithm 1.\rm          
\STATE Initialize the parameter vector $w[0]$.
\STATE Initialize the state $x[0]$.
\FOR{Discrete-time $k=0,...,K$}
\STATE Observe the state $x[k]$.
\STATE Determine the action $a[k]=\argmax_{a\in\mathcal{A}}Q(x[k],a|w[k])$.
\STATE Add the exploration noise $a[k]\leftarrow a[k]+\epsilon[k]$.
\STATE Execute the action $a[k]$ to the real system.
\STATE Receive the next state $x[k+1]$ and the reward $r[k]$ computed by Eq.\ (\ref{reward_assumed}).
\STATE Initialize the count of revising the update rate: $l\leftarrow0$.
\WHILE{True}    
\STATE Compute the next parameter vector $w[k+1]$:\\
$w[k+1]\leftarrow w[k]-\alpha2^{-l}\left( \frac{\partial \mathcal{L}(w[k])}{\partial w} +\eta \frac{\partial B(w[k])}{\partial w}\right)$.
\IF{All elements of $w[k+1]$ are positive}
\STATE \bf break \rm
\ENDIF
\STATE $l\leftarrow l+1$.
\ENDWHILE
\STATE Normalize the parameter vector $w[k+1]$:\\
 $w[k+1]\leftarrow\frac{1}{\sum_{j=1}^{N}w_{j}[k+1]}w[k+1]$.
\ENDFOR
\end{algorithmic}
\end{algorithm}

%%%%%%%%%%%%%%%%%%%%%%%%%%%%%%%%%%%%%%%%%%%%%%%%%%%%%%%%%%%%%%%%%%%%%%%%%%%%%%%%
\section{EXAMPLE}
\begin{figure}[htbp]
  \begin{center}
    \includegraphics[clip,width=8.0cm]{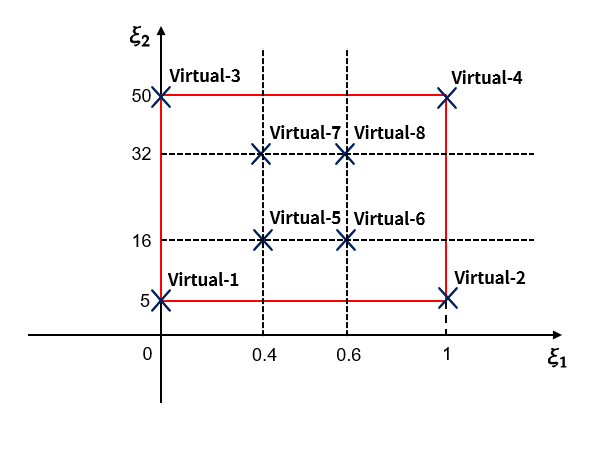}
    \caption{Parameter region where the system parameter vector of the real system lies. Parameters of virtual systems are denoted by cross marks.}
    \label{parameter_region}
  \end{center}
\end{figure}

We consider the following discrete-time system.
\begin{eqnarray}
&&\left[
    \begin{array}{ccc}
      x_{1}[k+1] \\
      x_{2}[k+1]
    \end{array}
  \right]\nonumber\\&&=
  \left[
    \begin{array}{ccc}
      x_{1}[k]+ d x_{2}[k] \\
      x_{2}[k] + d (g\sin(x_{1}[k])-\xi_{1}x_{2}[k]+\xi_{2}a[k])
    \end{array}
  \right],\nonumber\\
  \label{example_dynamics}
  \label{example}
\end{eqnarray}
where $g=9.81$ and $d=2^{-4}$. The state set and action set are $\mathcal{X}=\mathbb{R}^{2}$ and $\mathcal{A}=[-1,1]$, respectively. We assume an uncertain parameter vector of the real system $\xi =[\xi_{1}\ \xi_{2}]^{\text{T}}$ lies in a region $\Xi=\{(\xi_{1},\xi_{2})|\ 0\le \xi_{1}\le1,\ 5\le \xi_{1}\le50\}$. We prepare the following virtual systems as shown in Fig.\ \ref{parameter_region}:

\bf virtual system-1 \rm  $\xi^{(1)}=(\xi_{1}^{(1)},\xi_{2}^{(1)})=(0.0, 5.0)$,

\bf virtual system-2 \rm  $\xi^{(2)}=(\xi_{1}^{(2)}, \xi_{2}^{(2)})=(1.0, 5.0)$, 

\bf virtual system-3 \rm  $\xi^{(3)}=(\xi_{1}^{(3)}, \xi_{2}^{(3)})=(0.0, 50.0)$,

\bf virtual system-4 \rm  $\xi^{(4)}=(\xi_{1}^{(4)}, \xi_{2}^{(4)})=(1.0, 50.0)$,

\bf virtual system-5 \rm  $\xi^{(5)}=(\xi_{1}^{(5)}, \xi_{2}^{(5)})=(0.4, 16.0)$,

\bf virtual system-6 \rm  $\xi^{(6)}=(\xi_{1}^{(6)}, \xi_{2}^{(6)})=(0.6, 16.0)$,

\bf virtual system-7 \rm  $\xi^{(7)}=(\xi_{1}^{(7)}, \xi_{2}^{(7)})=(0.4, 32.0)$,

\bf virtual system-8 \rm  $\xi^{(8)}=(\xi_{1}^{(8)}, \xi_{2}^{(8)})=(0.6, 32.0)$.

We use    
\begin{eqnarray}
R_{1}=\left[
    \begin{array}{ccc}
      1.0 & 0 \\
      0 & 0.1
    \end{array}
  \right],\ \ \ R_{2}=10.0\nonumber
\end{eqnarray}
as positive definite matrices of the reward function (\ref{reward_assumed}). Let the origin be a target state $x^{*}=0$. Note that the origin is a fixed point of the system (\ref{example_dynamics}).
 
\begin{figure*}[htbp]
  \begin{center}
    \begin{tabular}{cc}

      % 1
      \begin{minipage}{0.3\hsize}
        \begin{center}
          \includegraphics[width=6.0cm]{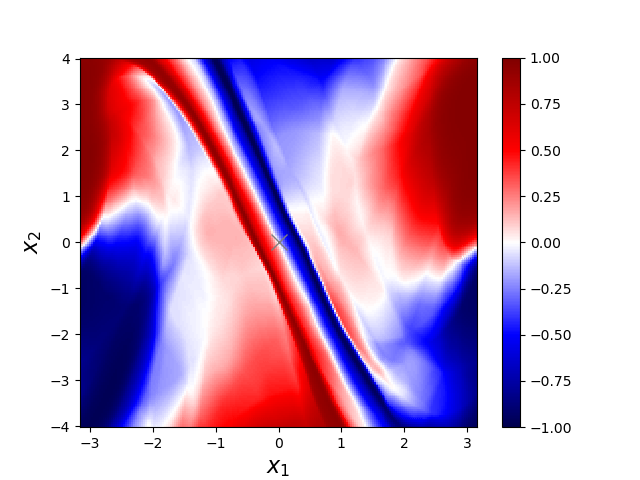}
        \end{center}
          \subcaption{Control policy $\mu_{1}^{*}$.}\label{mu_1}
      \end{minipage}
	%\hfill
      % 2
      \begin{minipage}{0.3\hsize}
        \begin{center}
          \includegraphics[width=6.0cm]{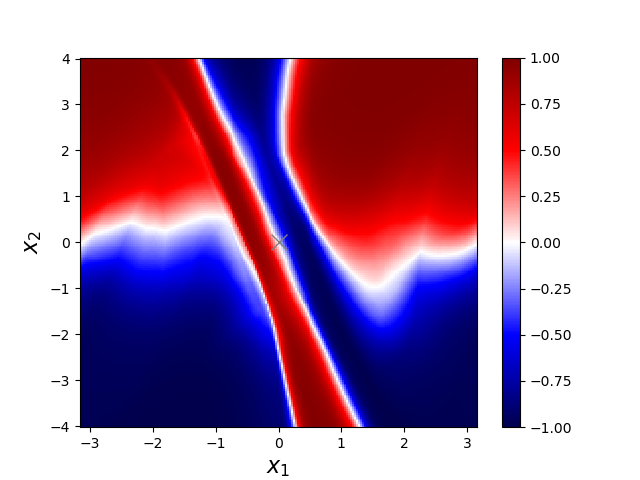}
        \end{center}
         \subcaption{Control policy $\mu_{2}^{*}$.}\label{mu_2}
      \end{minipage}\\
      %\hfill

      % 3
      \begin{minipage}{0.3\hsize}
        \begin{center}
          \includegraphics[width=6.0cm]{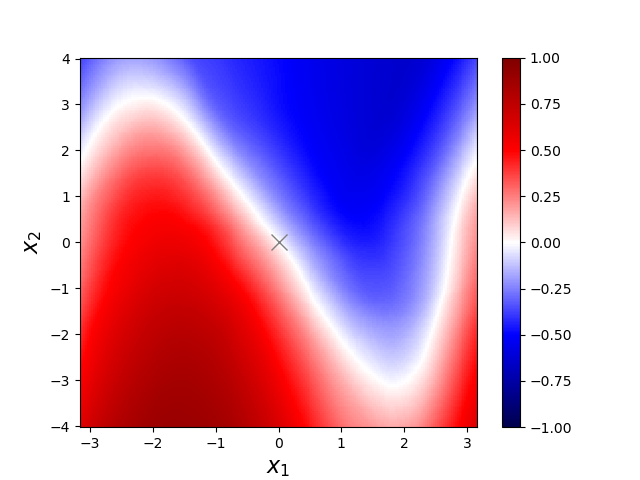}
        \end{center}
         \subcaption{Control policy $\mu_{3}^{*}$.}\label{mu_3}
      \end{minipage}
      
      % 4
      \begin{minipage}{0.3\hsize}
        \begin{center}
          \includegraphics[width=6.0cm]{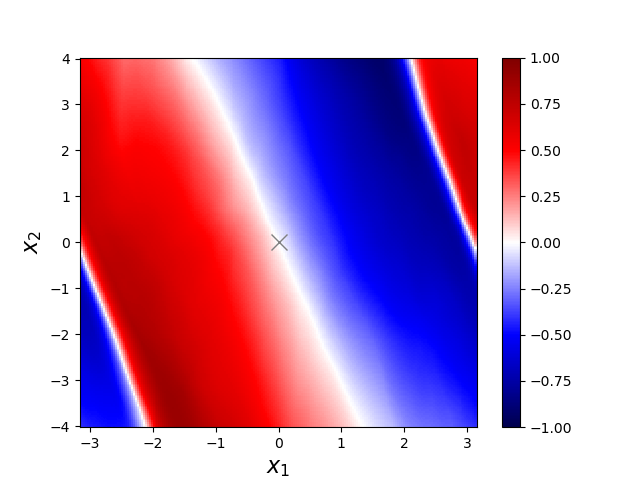}
        \end{center}
         \subcaption{Control policy $\mu_{4}^{*}$.}\label{mu_4}
      \end{minipage}
      
       % 5
      \begin{minipage}{0.3\hsize}
        \begin{center}
          \includegraphics[width=6.0cm]{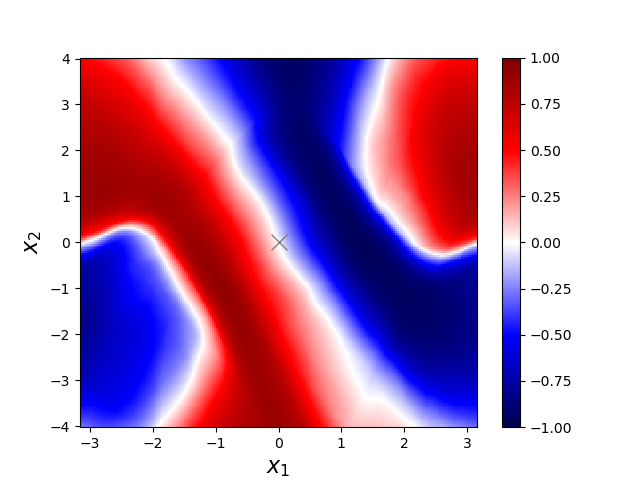}
        \end{center}
         \subcaption{Control policy $\mu_{5}^{*}$.}\label{mu_5}
      \end{minipage}\\
	%\hfill
      % 6
      \begin{minipage}{0.3\hsize}
        \begin{center}
          \includegraphics[width=6.0cm]{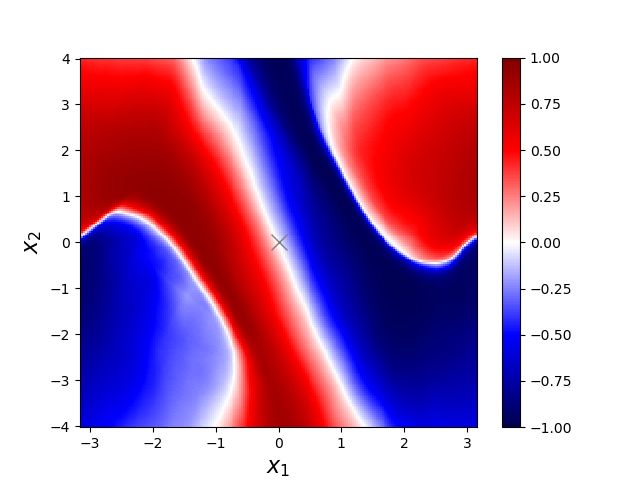}
        \end{center}
         \subcaption{Control policy $\mu_{6}^{*}$.}\label{mu_6}
      \end{minipage}
      %\hfill

      % 7
      \begin{minipage}{0.3\hsize}
        \begin{center}
          \includegraphics[width=6.0cm]{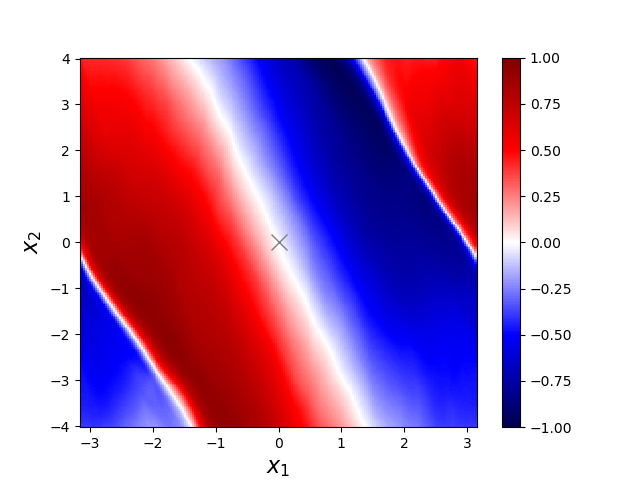}
        \end{center}
         \subcaption{Control policy $\mu_{7}^{*}$.}\label{mu_7}
      \end{minipage}
      
      % 8
      \begin{minipage}{0.3\hsize}
        \begin{center}
          \includegraphics[width=6.0cm]{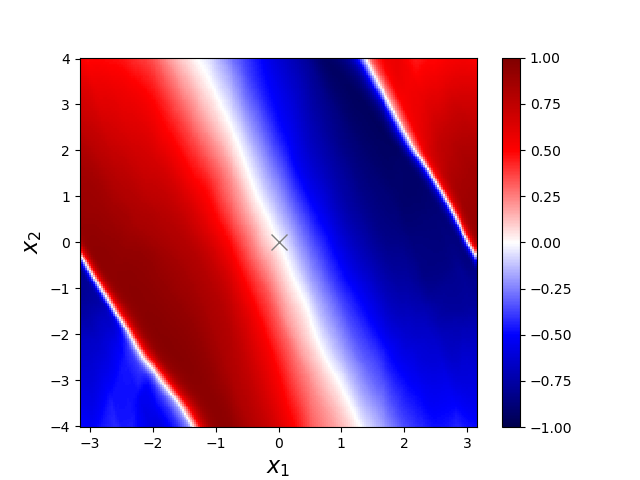}
        \end{center}
         \subcaption{Control policy $\mu_{8}^{*}$.}\label{mu_8}
      \end{minipage}

    \end{tabular}
    
    %\vskip20mm
    \caption{Illustrations of policies learned for virtual systems. Although these characteristics are different from each other, they determine close to 0 as the optimal action around the target state that is denoted by the cross mark.}
    \label{policy}
  \end{center}
\end{figure*}

We use the same DNN architecture to learn optimal Q-functions for all virtual systems. The DNN has four hidden layers, where all hidden layers have 128 units and all layers are fully connected layers. Activation functions are ReLU except for output layers. In regards to activation functions of output layers, we use hyperbolic tangent functions for units of optimal actions $\mu(\cdot|\theta^{\mu})$ and linear functions for the other units, respectively. The size of the replay buffer $\mathcal{B}$ is $1.0\times10^{6}$ and the size of the minibatch is $I=128$. These parameter vectors of DNNs are updated by Adam \cite{Adam}. In these simulations, the learning rate for \bf virtual system-1 \rm is $5.0\times10^{-4}$, these learning rates for \bf virtual system-2,\ 5,\ 6 \rm are $5.0\times10^{-5}$, and these learning rates for \bf virtual system-3,\ 4,\ 7,\ 8 \rm are $1.0\times10^{-4}$. The update rate of the target networks is $\tau=0.005$. The discounted factor is $\gamma=0.99$. We use the following \it Ornstein Uhlenbeck process \rm \cite{Ou_noise} to generate exploration noises $\epsilon^{\text{OU}}[k]$.
\begin{eqnarray}
\epsilon^{\text{OU}}[k+1]&=&\epsilon^{\text{OU}}[k] + p_{1}(p_{2}-\epsilon^{\text{OU}}[k])+p_{3}\epsilon', \nonumber\\
\epsilon^{\text{OU}}[0]&=&0,\nonumber
\end{eqnarray}
where $(p_{1},p_{2},p_{3})=(0.15,0.0,0.3)$ and $\epsilon'$ is a noise generated by the standard normal distribution. The learned Q-function and policy for the virtual system with $\xi^{(j)}$ are denoted by $Q_{j}^{*}$ and $\mu_{j}^{*}$, respectively. These policies learned for virtual systems are shown in Fig.\ \ref{policy}. Although their characteristics are different from each other, all actions determined by them are close to 0 around the target state since it is a fixed point of the system. 

%%%%%%%%%%%%%%%%%%%%%%%%%%%%%%%%%%%%%%%%%%%%%%%
%%%%%%%%%%%%%%%%%%%%%%%%%%%%%%%%%%%%%%%%%%%%%%%
%%%%%%%%%%%%%%%%%%%%%%%%%%%%%%%%%%%%%%%%%%%%%%%
\begin{figure*}[htbp]
  \begin{center}
    \begin{tabular}{cc}

      % 1
      \begin{minipage}{0.3\hsize}
        \begin{center}
          \includegraphics[width=6.0cm]{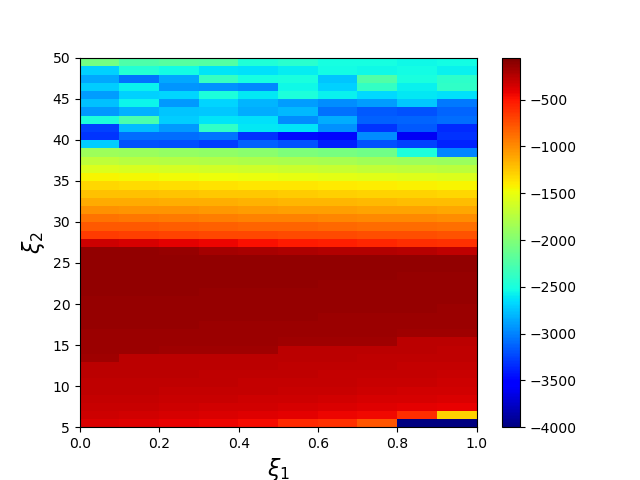}
        \end{center}
          \subcaption{Control policy $\mu_{1}^{*}$.}\label{robust_1}
      \end{minipage}
	%\hfill
      % 2
      \begin{minipage}{0.3\hsize}
        \begin{center}
          \includegraphics[width=6.0cm]{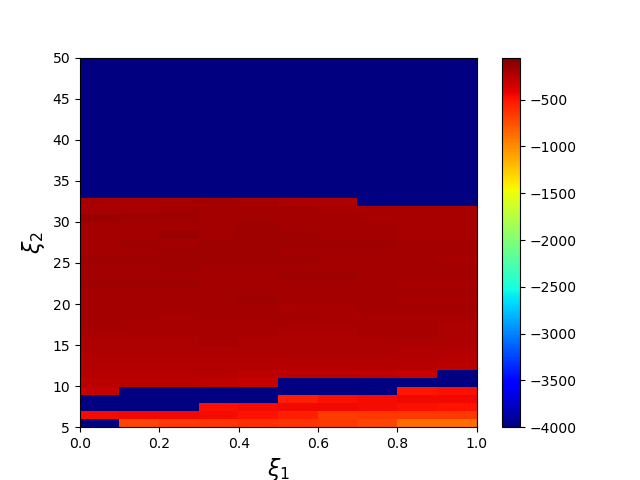}
        \end{center}
          \subcaption{Control policy $\mu_{2}^{*}$.}\label{robust_2}
      \end{minipage}\\
      %\hfill

      % 3
      \begin{minipage}{0.3\hsize}
        \begin{center}
          \includegraphics[width=6.0cm]{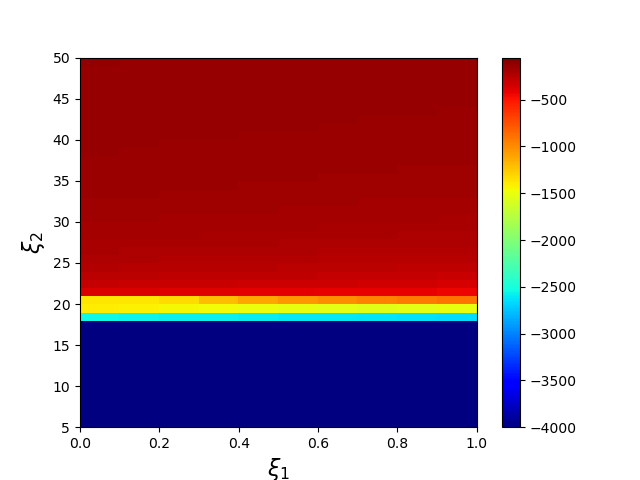}
        \end{center}
          \subcaption{Control policy $\mu_{3}^{*}$.}\label{robust_3}
      \end{minipage}
      
      % 4
      \begin{minipage}{0.3\hsize}
        \begin{center}
          \includegraphics[width=6.0cm]{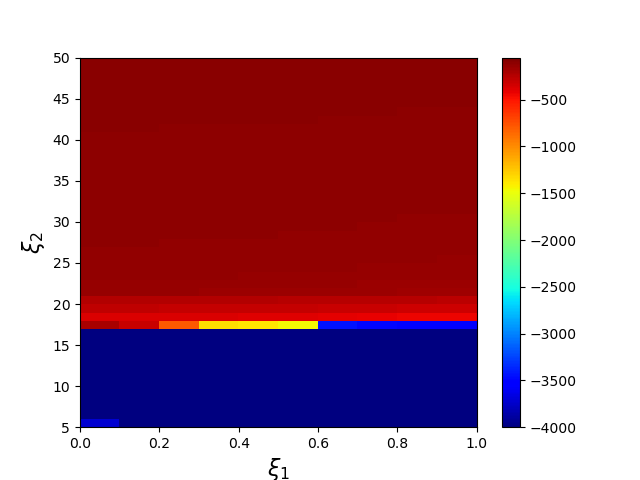}
        \end{center}
          \subcaption{Control policy $\mu_{4}^{*}$.}\label{robust_4}
      \end{minipage}
      
       % 5
      \begin{minipage}{0.3\hsize}
        \begin{center}
          \includegraphics[width=6.0cm]{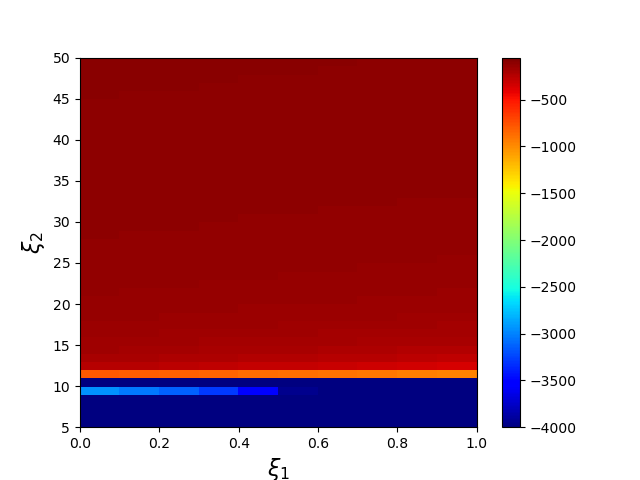}
        \end{center}
          \subcaption{Control policy $\mu_{5}^{*}$.}\label{robust_5}
      \end{minipage}\\
	%\hfill
      % 6
      \begin{minipage}{0.3\hsize}
        \begin{center}
          \includegraphics[width=6.0cm]{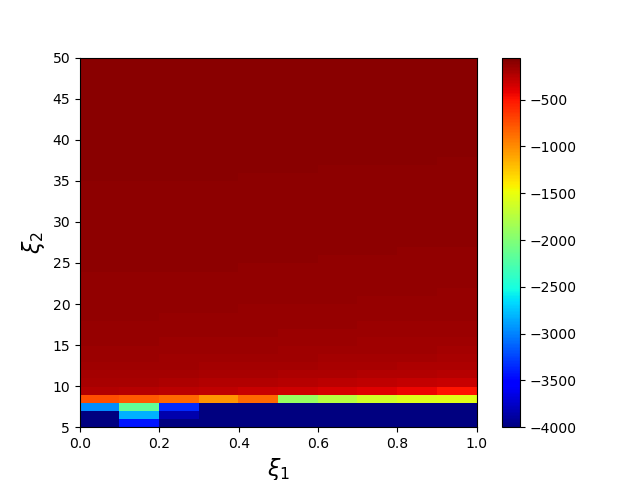}
        \end{center}
          \subcaption{Control policy $\mu_{6}^{*}$.}\label{robust_6}
      \end{minipage}
      %\hfill

      % 7
      \begin{minipage}{0.3\hsize}
        \begin{center}
          \includegraphics[width=6.0cm]{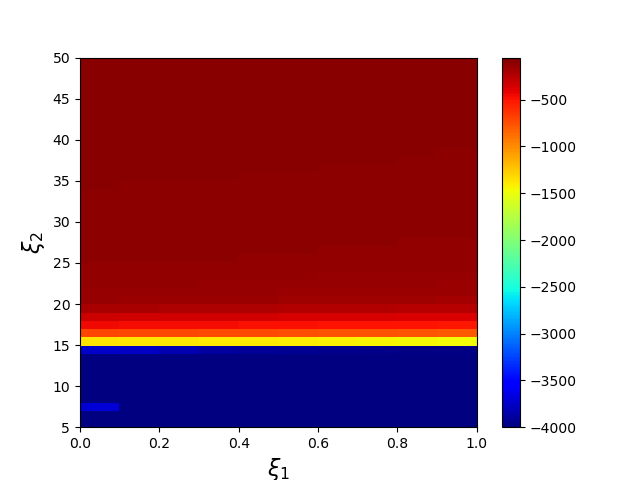}
        \end{center}
          \subcaption{Control policy $\mu_{7}^{*}$.}\label{robust_7}
      \end{minipage}
      
      % 8
      \begin{minipage}{0.3\hsize}
        \begin{center}
          \includegraphics[width=6.0cm]{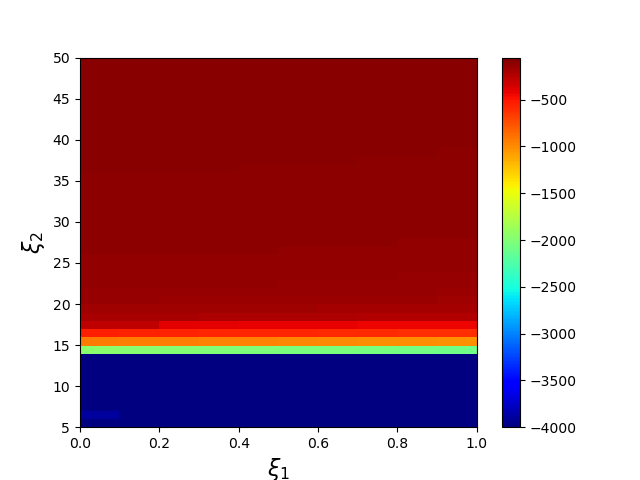}
        \end{center}
          \subcaption{Control policy $\mu_{8}^{*}$.}\label{robust_8}
      \end{minipage}

    \end{tabular}
    
    %\vskip25mm

    \caption{Scores of pre-trained policies for real systems with $\xi=(\xi_{1},\xi_{2})\in\Xi_{\text{plot}}$. Each grid shows the score of the pre-trained policy $\mu_{j}^{*},\ j=1,2,...,N,$ for the system with $\xi$. }
    \label{Robust_map}
  \end{center}
\end{figure*}
%%%%%%%%%%%%%%%%%%%%%%%%%%%%%%%%%%%%%%%%%%%%%%%

We define a \it score \rm 
\begin{eqnarray}
G(\mu|\xi)=\sum_{k=0}^{1000}R(x[k],\mu(x[k]))\label{robustness}
\end{eqnarray}
as the index of the policy's performance for the system with $\xi\in\Xi$, where
\begin{eqnarray}
x[k+1]=f(x[k],\mu(x[k])|\xi),\ x[0]=[\pi\ 0]^{\text{T}}.\nonumber
\end{eqnarray}
In the following simulations, if the agent obtains a score that is smaller than $-2000$, we consider that the agent's policy $\mu$ does not perform well for the real system with $\xi$. In order to show the performance of the policy $\mu_{j}^{*}$, we plot scores as shown in Fig.\ \ref{Robust_map}. We show scores for real systems with $\xi=(\xi_{1},\xi_{2})\in\Xi_{\text{plot}}$, where $\Xi_{\text{plot}}=\{0.05,0.15,...,0.95\}\times\{5.5, 6.5,...,49.5\}$. The policy $\mu_{j}^{*}$ performs well for real systems whose system parameter vectors are close to $\xi^{(j)}$. However, it is shown that the policy learned with the simulator does not perform well for the real system if we have an identification error. Thus, we apply our proposed method to this problem using these Q-functions learned with the simulator. The barrier function (\ref{barrier_function}) is given by 
\begin{eqnarray}
B(w)=-\sum_{j=1}^{N} \log (w_{j}+\epsilon_{w}),\ w\in\text{int}\mathcal{W}, \label{barrier_function_w}
\end{eqnarray}
where $\mathcal{W}=\{w|\ \forall j\in\{1,2,...,N\},\ w_{j}+\epsilon_{w}>0\}$. $B(w)$ diverges as the parameter $w$ approaches the boundary $\partial \mathcal{W}$. It is known as a \it log barrier function\rm. We set $\eta=1.0\times10^{-7}$ and $\epsilon_{w}=1.0\times10^{-9}$, respectively. 

All experiments were run on a computer with an Intel(R) Core(TM) i7-10700 @ 2.9GHz processor and 32GB of memory. We conducted all experiments with Python. 

\subsection{Choice of Basis Functions}
In this section, we discuss the relationship between the choice of basis functions and the performances of the stabilization. The exploration noises are generated by 
\begin{eqnarray}
\epsilon[k] =0.1 \frac{\max(400-k,0)}{400}\epsilon',\label{noise_case1}
\end{eqnarray}
where noises $\epsilon'$ is generated by the standard normalized distribution. We do not add exploration noises after the 400th step. The initial state is $[\pi\ 0]^{\text{T}}$. The learning rate is $\alpha=5.0\times10^{-5}$. The max step is $K=1000$. We set elements of the initial parameter vector $w_{j}=\frac{1}{N},\ j=1,2,...,N$. 

\begin{figure*}[htbp]
  \begin{center}
    \begin{tabular}{cc}

      % 1
      \begin{minipage}{0.30\hsize}
        \begin{center}
          \includegraphics[width=5.5cm]{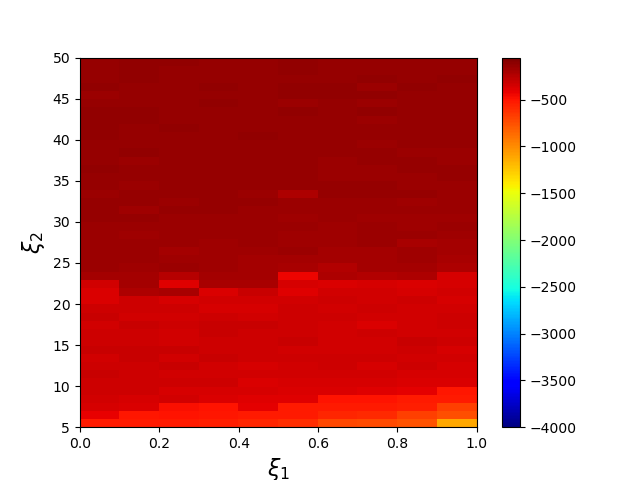}
        \end{center}
        \subcaption{\bf Case-1\rm}\label{case1}
      \end{minipage}
	%\hfill
      % 2
      \begin{minipage}{0.30\hsize}
        \begin{center}
          \includegraphics[width=5.5cm]{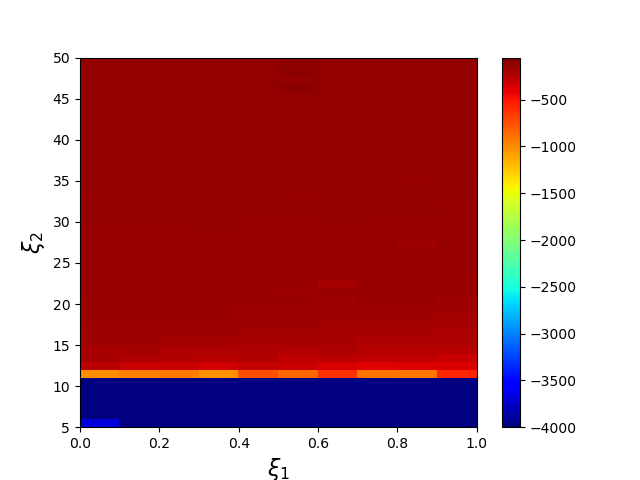}
        \end{center}
        \subcaption{\bf Case-2\rm}\label{case2}
      \end{minipage}\\
      %\hfill

      % 3
      \begin{minipage}{0.30\hsize}
        \begin{center}
          \includegraphics[width=5.5cm]{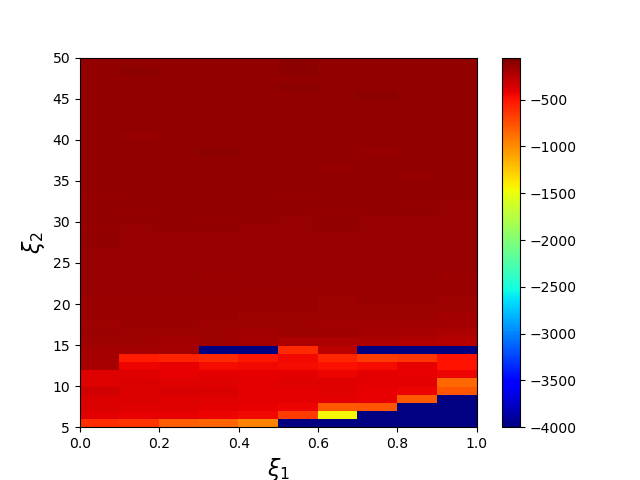}
        \end{center}
        \subcaption{\bf Case-3\rm}\label{case3}
      \end{minipage}
      
      % 4
      \begin{minipage}{0.30\hsize}
        \begin{center}
          \includegraphics[width=5.5cm]{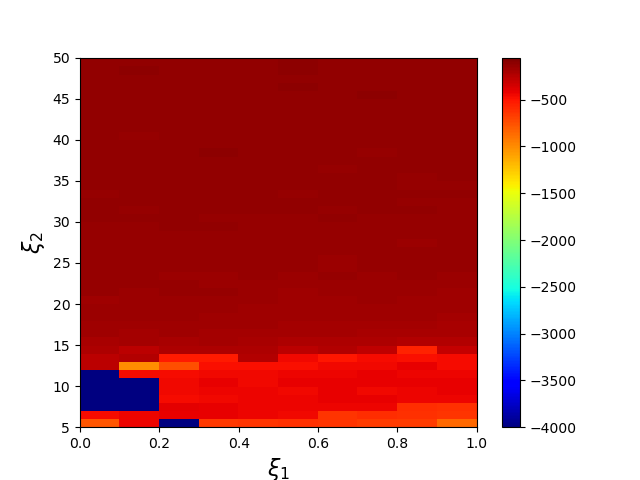}
        \end{center}
        \subcaption{\bf Case-4\rm}\label{case4}
      \end{minipage}
      
       % 5
      \begin{minipage}{0.30\hsize}
        \begin{center}
          \includegraphics[width=5.5cm]{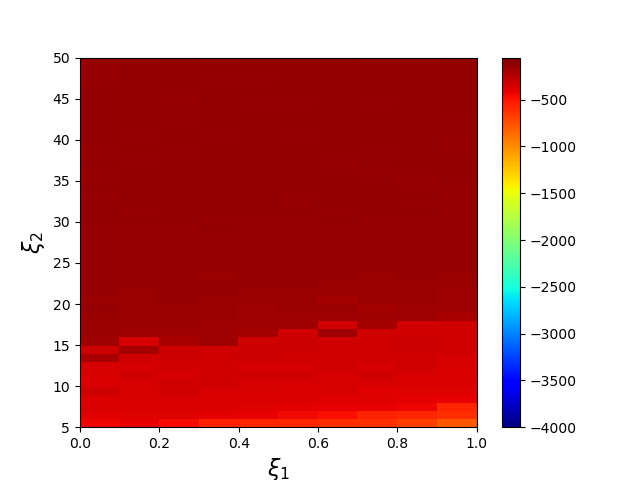}
        \end{center}
        \subcaption{\bf Case-5\rm}\label{case5}
      \end{minipage}
	%\hfill

    \end{tabular}
    
    %\vskip25mm

    \caption{Scores of policies learned by our proposed method online for real systems with $\xi=(\xi_{1},\xi_{2})\in\Xi_{\text{plot}}$. Each grid shows the score $G(\mu(\cdot|w)|\xi)$ for the real system with $\xi$.}
    \label{result_IV}
  \end{center}
\end{figure*}

\begin{figure}
  \begin{center}
    \includegraphics[clip,width=8.5cm]{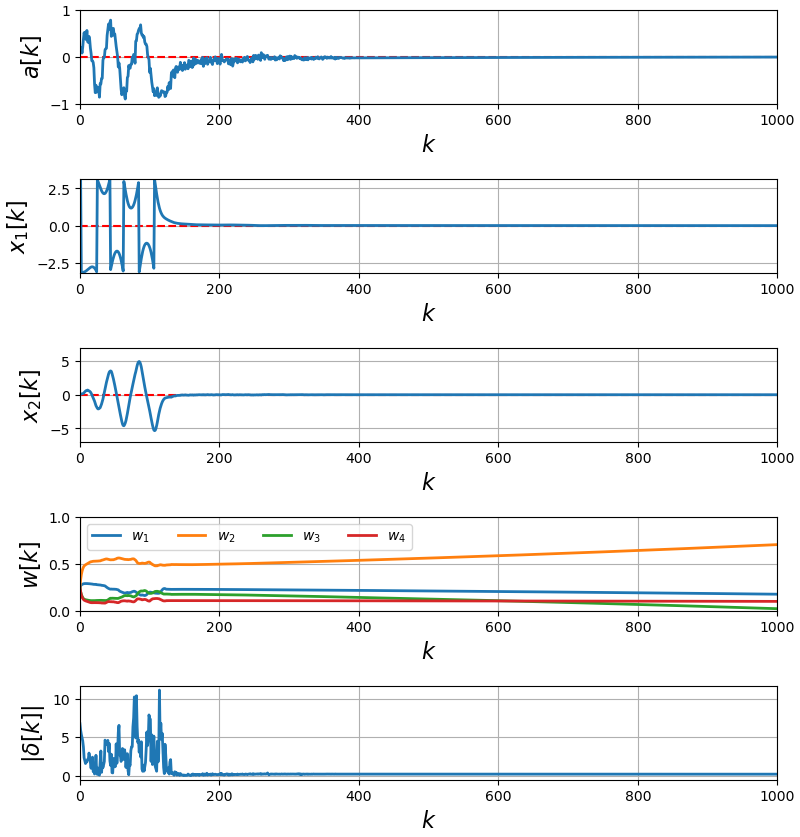}
    \caption{The time response of the real system with $(\xi_{1},\xi_{2})=(0.95, 5.5)$ controlled by the agent that learns the parameter vector $w=[w_{1}\ w_{2}\ w_{3}\ w_{4}]^{\text{T}}$ using our proposed method online, where $w_{j}$ is the weight of the optimal Q-function $Q_{j}^{*}$. $|\delta[k]|$ is the TD-error at the step $k$. }
    \label{result_IV_A_N4_case1_TR}
  \end{center}
\end{figure}

\begin{figure}
  \begin{center}
    \includegraphics[clip,width=8.5cm]{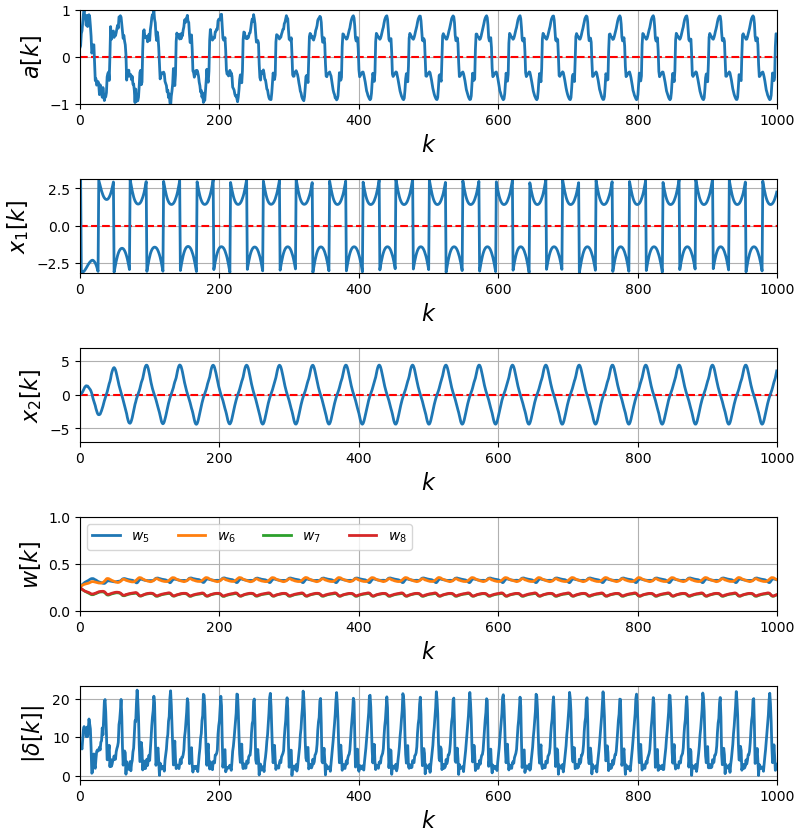}
    \caption{The time response of the real system with $(\xi_{1},\xi_{2})=(0.95, 5.5)$ controlled by the agent that learns the parameter vector $w=[w_{5}\ w_{6}\ w_{7}\ w_{8}]^{\text{T}}$ using our proposed method, where $w_{j}$ is the weight of the optimal Q-function $Q_{j}^{*}$. $|\delta[k]|$ is the TD-error at the step $k$. }
    \label{result_IV_A_N4_case2_TR}
  \end{center}
\end{figure}

At first, we assume that $N=4$ and consider five cases summarized in TABLE \ref{case_table} for the choice of basis functions. For each case, Fig.\ \ref{result_IV} shows scores of policies learned by our proposed algorithm online for real systems with $\xi\in\Xi_{\text{plot}}$. The scores for \bf Case-1 \rm are shown in Fig.\ \ref{result_IV}\subref{case1}. It is shown that the agent with $\{Q_{1}^{*},Q_{2}^{*},Q_{3}^{*},Q_{4}^{*}\}$ learns policies that perform well for real systems with $(\xi_{1},\xi_{2})\in\Xi_{\text{plot}}$. Additionally, we show the time response for the real system with $(\xi_{1},\xi_{2})=(0.95,5.5)$ in Fig. \ref{result_IV_A_N4_case1_TR}. It is shown that the agent stabilizes the real system around the target state. The scores for \bf Case-2 \rm are shown in Fig.\ \ref{result_IV}\subref{case2}. The policy learned with $\{Q_{5}^{*},Q_{6}^{*},Q_{7}^{*},Q_{8}^{*}\}$ does not perform well for a system if its system parameter $\xi_{2}$ is smaller than 10. The time response of the real system with $(\xi_{1},\xi_{2})=(0.95,5.5)$ is shown in Fig.\ \ref{result_IV_A_N4_case2_TR}. It is shown that the agent does not stabilize the real system. We consider \bf Case-3 \rm and \bf Case-4 \rm to confirm that $Q_{1}^{*}$ and $Q_{2}^{*}$ are necessary to control a real system whose system parameter $\xi_{2}$ is small. The scores for \bf Case-3 \rm and \bf Case-4 \rm are shown in Figs.\ \ref{result_IV}\subref{case3} and \ref{result_IV}\subref{case4}, respectively. It is shown that the learned policies do not perform well for some real systems. Consequently, we need both $Q_{1}^{*}$ and $Q_{2}^{*}$ as basis functions. In \bf Case-5 \rm where we use both $Q_{1}^{*}$ and $Q_{2}^{*}$, the agent learns its policy that performs well for systems with $\xi\in\Xi_{\text{plot}}$ as shown in Fig. \ref{result_IV}\subref{case5}. 
 
\begin{table}[htb]
  \begin{center}
    \caption{Choice of basis functions for the Q-function}
    \begin{tabular}{|c||c|} \hline
    Case number & Choice of basis functions \\ \hline \hline
      \bf Case-1 \rm& $\{Q_{1}^{*},\ Q_{2}^{*},\ Q_{3}^{*},\ Q_{4}^{*}\} $\\  
      \bf Case-2 \rm&  $\{Q_{5}^{*},\ Q_{6}^{*},\ Q_{7}^{*},\ Q_{8}^{*}\}$ \\ 
      \bf Case-3 \rm& $\{Q_{1}^{*},\ Q_{6}^{*},\ Q_{7}^{*},\ Q_{8}^{*}\}$\\ 
      \bf Case-4 \rm& $\{Q_{5}^{*},\ Q_{2}^{*},\ Q_{7}^{*},\ Q_{8}^{*}\}$\\ 
       \bf Case-5 \rm& $\{Q_{1}^{*},\ Q_{2}^{*},\ Q_{7}^{*},\ Q_{8}^{*}\}$ \\ \hline
    \end{tabular}
    \label{case_table}
  \end{center}
\end{table}

%%%%%%%%%%%%%%%%%%%%%%%%%%%%%%%%%%%%%%%%%%%%%%%
%%%%%%%%%%%%%%%%%%%%%%%%%%%%%%%%%%%%%%%%%%%%%%%

%%%%%%%%%%%%%%%%%%%%%%%%%%%%%%%%%%%%%%%%%%%%%%%

%%%%%%%%%%%%%%%%%%%%%%%%%%%%%%%%%%%%%%%%%%%%%%%
%%%%%%%%%%%%%%%%%%%%%%%%%%%%%%%%%%%%%%%%%%%%%%%  

\begin{figure*}[htbp]
  \begin{center}
    \begin{tabular}{c}

      % 1
      \begin{minipage}{0.30\hsize}
        \begin{center}
          \includegraphics[width=5.5cm]{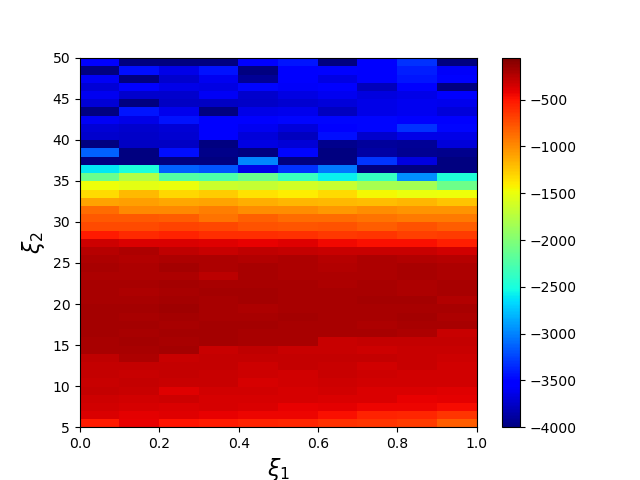}
        \end{center}
        \subcaption{$N=2$ $(\{Q_{1}^{*},Q_{2}^{*}\})$}\label{N2}
      \end{minipage}
	%\hfill
      % 2
      \begin{minipage}{0.30\hsize}
        \begin{center}
          \includegraphics[width=5.5cm]{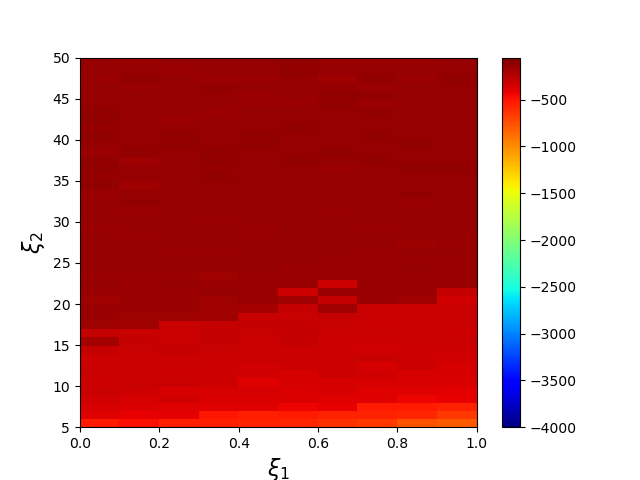}
        \end{center}
        \subcaption{$N=3$ $(\{Q_{1}^{*},Q_{2}^{*},Q_{4}^{*}\})$}\label{N3}
      \end{minipage}
      %\hfill

      % 3
      \begin{minipage}{0.30\hsize}
        \begin{center}
          \includegraphics[width=5.5cm]{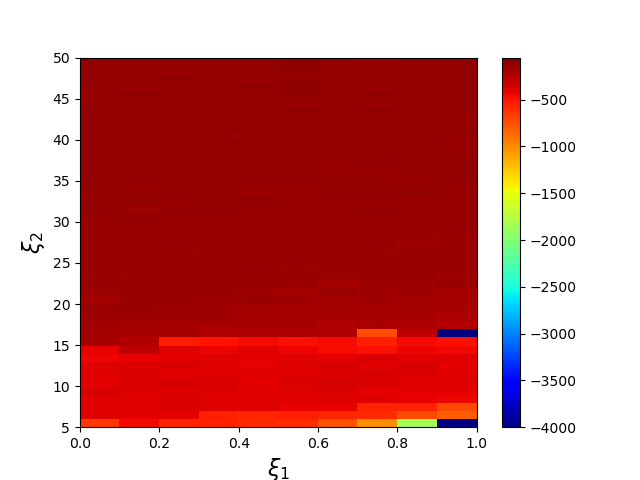}
        \end{center}
        \subcaption{$N=8$ ($\{Q_{1}^{*},Q_{2}^{*},...,Q_{8}^{*}\}$)}\label{N8}
      \end{minipage}
    \end{tabular}
    
   \caption{Scores of policies learned by our proposed method online for systems with $\xi=(\xi_{1},\xi_{2})\in\Xi_{\text{plot}}$. We consider three cases: $\{Q_{1}^{*},Q_{2}^{*}\}$, $\{Q_{1}^{*},Q_{2}^{*},Q_{4}^{*}\}$, and $\{Q_{1}^{*},Q_{2}^{*},...,Q_{8}^{*}\}$. Each grid shows $G(\mu(\cdot|w)|\xi)$ for the real system with $\xi=(\xi_{1},\xi_{2})\in\Xi_{\text{plot}}$.}
    \label{result_IV_B}
  \end{center}
\end{figure*}

Next, we consider how many basis functions we should use. These learned policies $\mu_{3}^{*}$, $\mu_{4}^{*}$, $\mu_{5}^{*}$, $\mu_{6}^{*}$, $\mu_{7}^{*}$, and $\mu_{8}^{*}$ perform well for real systems if $\xi_{2}$ are larger than 35, as shown in Fig.\ \ref{Robust_map}. If we choose at least one of $\{Q_{3}^{*},Q_{4}^{*},...,Q_{8}^{*}\}$ as basis functions, the agent may learn the policy that performs well for such real systems. Then, we consider the case where we choose $\{Q_{1}^{*},Q_{2}^{*},Q_{4}^{*}\}$ as basis functions. Scores of policies learned for systems with $\xi\in\Xi_{\text{plot}}$ are shown in Fig.\ \ref{result_IV_B}\subref{N3}. It is shown that the agent learns policies that perform well for systems with $\xi\in\Xi_{\text{plot}}$. On the other hand, if we use $Q_{1}^{*}$ and $Q_{2}^{*}$ only, the agent does not learn policies that perform well for real systems if $\xi_{2}$ is larger than 35, as shown in Fig.\ \ref{result_IV_B}\subref{N2}. We should choose $Q_{1}^{*}$, $Q_{2}^{*}$, and at least one of $\{Q_{3}^{*},Q_{4}^{*},...,Q_{8}^{*}\}$. Moreover, we consider the case where we choose all optimal Q-functions learned for virtual systems as basis functions. Then, the representation of the Q-function is redundant. Scores of policies learned for real systems with $\xi\in\Xi_{\text{plot}}$ is shown in Fig.\ \ref{result_IV_B}\subref{N8}. Although the agent performs well for most real systems, it does not learn policies that perform well for the real systems with $(\xi_{1},\xi_{2})=(0.95,5.5)$ and $(0.95,16.5)$. If we choose basis functions redundantly, the agent may not learn the policy for the real system. 

%%%%%%%%%%%%%%%%%%%%%%%%%%%%%%%%%%%%%%%%%%%%%%%
%%%%%%%%%%%%%%%%%%%%%%%%%%%%%%%%%%%%%%%%%%%%%%%

%%%%%%%%%%%%%%%%%%%%%%%%%%%%%%%%%%%%%%%%%%%%%%%
In our proposed method, through interactions with the real system, the agent learns the approximated linear Q-function whose basis functions are optimal Q-functions learned for virtual systems. To achieve good performances for a set of system parameter vectors as large as possible, we choose a set of Q-functions such that system parameters sets stabilized by the Q-functions are complementary to each other. Moreover, it is desirable to reduce the number of basis functions as much as possible. 

\subsection{Adaptivity for The Varying System}
We show that our proposed method can be applied to a real system whose system parameter vector varies slowly. In the following, we choose $\{Q_{1}^{*},Q_{2}^{*},Q_{4}^{*}\}$ as basis functions. The initial parameter vector of the Q-function is $[1/3\ 1/3\ 1/3]^{\text{T}}$. We add exploration noises to actions if $\left|\left|x\right|\right|_{2}\ge0.05$. These noises are generated by the standard normalized distribution, where we multiply these noises by 0.1. The initial state is $[\pi\ 0]^{\text{T}}$. The learning rate is $\alpha=5.0\times10^{-5}$. The max step is $K=1000$.

\begin{figure*}[htbp]
  \begin{center}
    \begin{tabular}{c}

      % 1
      \begin{minipage}{0.50\hsize}
        \begin{center}
          \includegraphics[width=8.5cm]{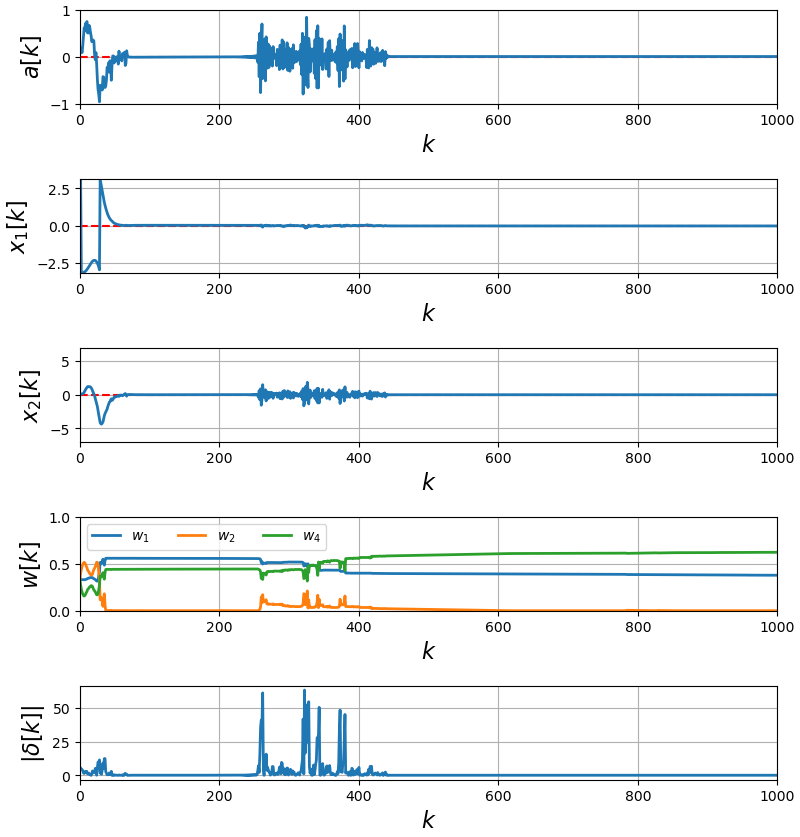}
        \end{center}
        \subcaption{Variation of $\xi_{2}$ from $5.0$ to $50.0$.}\label{Change_env_1}
      \end{minipage}
	%\hfill
      % 2
      \begin{minipage}{0.50\hsize}
        \begin{center}
          \includegraphics[width=8.5cm]{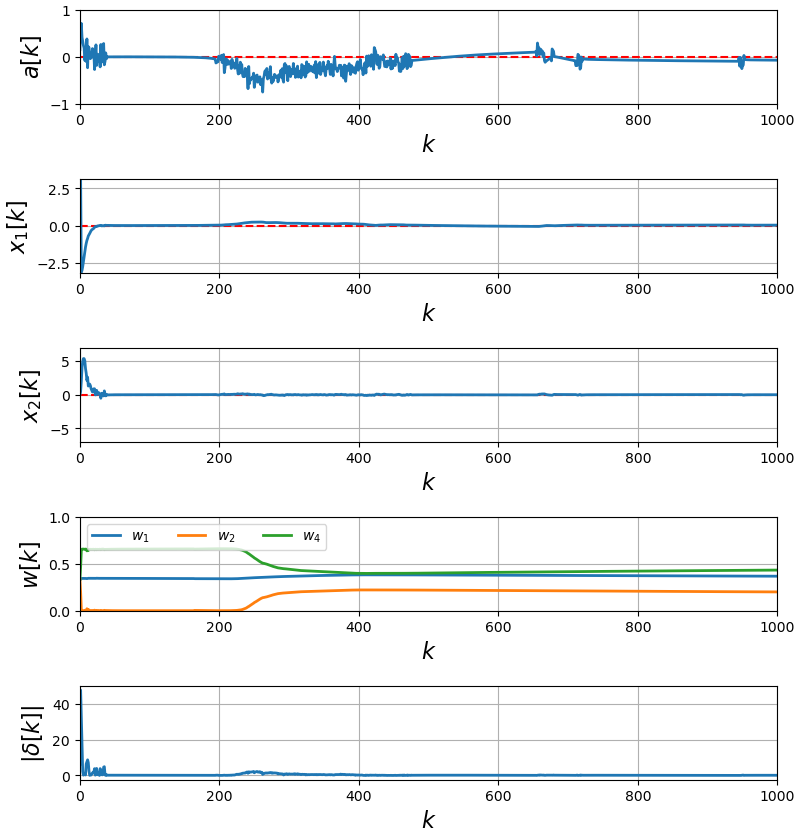}
        \end{center}
        \subcaption{Variation of $\xi_{2}$ from $50.0$ to $5.0$.}\label{Change_env_3}
      \end{minipage}
      %\hfill
      
       \end{tabular}

   \caption{The time response of the system controlled by the agent that learns the parameter vector $w$ by our proposed method online. It is assumed that the system parameter $\xi_{2}$ varies slowly until $k=200$, where $\xi_{1}=1.0$.   }
    \label{result_IV_C}
  \end{center}
\end{figure*}

First, it is assumed that the system parameter $\xi_{2}$ increases from $5.0$ to $50.0$ gradually until $k=200$, where $\xi_{1}=1.0$. The time response of the real system is shown in Fig.\ \ref{result_IV_C}\subref{Change_env_1}. Although the Euclidean norm of the system's state become larger than 0.05 once after $k=200$, the agent adds the exploration noise to its action and learns the parameter vector $w$. 

Second, it is assumed that the system parameter $\xi_{2}$ decreases from $50.0$ to $5.0$ gradually until $k=200$, where $\xi_{1}=1.0$. The time response of the real system is shown in Fig.\ \ref{result_IV_C}\subref{Change_env_3}. Although the Euclidean norm of the system's state sometimes become larger than 0.05, the agent can control the real system to the target state again by learning $w$ online. 

By the above results, the agent can adapt the real system whose system parameter vector varies within the premised set $\Xi$ by learning the parameter vector $w$.

%%%%%%%%%%%%%%%%%%%%%%%%%%%%%%%%%%%%%%%%%%%%%%%

%%%%%%%%%%%%%%%%%%%%%%%%%%%%%%%%%%%%%%%%%%%%%%%%%%%%%%%%%%%%%%%%%%%%%%%%%%%%%%%%
\section{CONCLUSIONS AND FUTURE WORKS}
We proposed a practical Q-learning algorithm with pre-trained multiple optimal Q-functions. Our proposed method consists of two stages. At the first stage, we obtain optimal Q-functions for virtual systems using the continuous deep Q-learning algorithm. At the second stage, we represent the Q-function for the real system by the approximated linear function whose basis functions are optimal Q-functions learned at the first stage. The agent learns the parameter vector of the approximated linear Q-function through interactions with the real system online. By numerical simulations, we show that the agent can learn the parameter vector and stabilize the real system to the target state. Moreover, we show that the agent can adapt to variations of the system parameter vector.

Selecting virtual systems' parameter vectors from the premised set and their numbers automatically is interesting future work. It is also future work to consider the case where we cannot identify only parameter vectors but also parts of real systems' models.

\section*{Appendix}
From (\ref{optimal_u}), we have
\begin{eqnarray}
&&\frac{\partial}{\partial a}\sum_{j=1}^{N}w_{j}A_{j}^{*}(x,a|\theta^{A_{j}})=0\nonumber\\
&\Leftrightarrow& -\sum_{j=1}^{N}w_{j}P_{j}^{*}(x|\theta^{P_{j}})(a-\mu_{j}^{*}(x|\theta^{\mu_{j}}))=0\nonumber\\
&\Leftrightarrow& \sum_{j=1}^{N}w_{j}P_{j}^{*}(x|\theta^{P_{j}})a=\sum_{j=1}^{N}w_{j}P_{j}^{*}(x|\theta^{P_{j}})\mu_{j}^{*}(x|\theta^{\mu_{j}}). \nonumber \\
\label{Appendix_1}
\end{eqnarray}
Then, $\sum_{j=1}^{N}w_{j}P_{j}^{*}(x|\theta^{P_{j}})$ is a positive definite symmetric matrix because $w_{j}>0$ and parameter matrices of NAFs $P_{j}^{*}$ are positive definite symmetric matrices. Hence, there is the inverse matrix $(\sum_{j=1}^{N}w_{j}P_{j}^{*}(x|\theta^{P_{j}}))^{-1}$ and the solution $\hat{a}$ of Eq.\ (\ref{Appendix_1}) as follows:
\begin{eqnarray}
\hat{a}&=&\left(\sum_{m=1}^{N}P_{m}^{*}(x|\theta^{P_{m}})\right)^{-1}\sum_{j=1}^{N}w_{j}P_{j}^{*}(x|\theta^{P_{j}})\mu_{j}^{*}(x|\theta^{\mu_{j}})\nonumber\\
&=&\sum_{j=1}^{N}\left(\sum_{m=1}^{N}P_{m}^{*}(x|\theta^{P_{m}})\right)^{-1}w_{j}P_{j}^{*}(x|\theta^{P_{j}})\mu_{j}^{*}(x|\theta^{\mu_{j}})\nonumber\\
&=&\sum_{j=1}^{N}\tilde{w}_{j}(x)\mu_{j}^{*}(x|\theta^{\mu_{j}}).\label{Appendix_2}
\end{eqnarray}

\ifCLASSOPTIONcaptionsoff
  \newpage
\fi

\end{document}